\documentclass{article}

\usepackage{arxiv}

\usepackage[utf8]{inputenc} % allow utf-8 input
\usepackage[T1]{fontenc}    % use 8-bit T1 fonts
\usepackage{nicefrac}       % compact symbols for 1/2, etc.
\usepackage{microtype}      % microtypography
\usepackage[numbers,sort&compress]{natbib}
\usepackage{doi}

\usepackage{listings}
\usepackage{hyperref}
\usepackage{pifont}
\usepackage{colortbl}
\usepackage[dvipsnames]{xcolor}

\usepackage{booktabs}
\usepackage{multirow}
\usepackage{bm}
\usepackage{geometry}
\usepackage{pdflscape}
\usepackage{tcolorbox}
\usepackage{tikz}
\usepackage{pgfplots}
\pgfplotsset{compat=1.18}
\usepackage{xspace}
\usepackage{tablefootnote}
\usepackage{enumitem}
\usepackage{afterpage}
\usepackage{nicematrix}
\usepackage{physics}
\usepackage{algorithmic} 
\usepackage[ruled,linesnumbered]{algorithm2e}
\usepackage{graphicx}
\usepackage{ifthen}
\usepackage{amsmath,amssymb}
\usepackage{enumitem}
\usepackage{amsfonts}
\usepackage{bold-extra}
\usepackage{caption}
\usepackage{subcaption}
\usepackage{siunitx}

\newboolean{showcomments}
\setboolean{showcomments}{true}         
\ifthenelse{\boolean{showcomments}}
  {\newcommand{\nb}[2]{
  \fbox{\bfseries\sffamily\scriptsize#1}
     {\small$\blacktriangleright$\textit{\textcolor{red}{#2}}$\blacktriangleleft$}
   }
  }
  {\newcommand{\nb}[2]{}
   
  }
\newcommand\nargiz[1]{\nb{Nargiz}{#1}}
\newcommand\gianmarco[1]{\nb{Gianmarco}{#1}}
\newcommand\paolo[1]{\nb{Paolo}{#1}}

\newcommand{\changed}[1]{{\color{black}#1}}

\newcommand{\topomap}{\textsc{TopoMap}\@\xspace}

\NewTColorBox{custombox}{o}{
    colback=gray!10,
    colframe=gray!10,
    left=1.0mm,
    right=1.0mm,
    top=1.0mm,
    bottom=1.0mm,
    fonttitle=\bfseries,
    arc=0mm,
    leftrule=0mm,
    rightrule=0mm,
    toprule=0mm,
    bottomrule=0mm,
    notitle,
    before=\par\medskip\noindent,
    IfValueTF={#1}{before upper={\textbf{#1: } }}{},
    parbox=false,
}

\title{TopoMap: A Feature-based Discriminator of the Topographical Regions in the Test Input Space}

\author{ \href{https://orcid.org/0009-0004-1546-4833}{\includegraphics[scale=0.06]{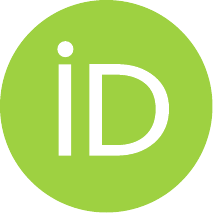}\hspace{1mm}Gianmarco De Vita} \\
	   Università della Svizzera italiana\\
            Lugano, Switzerland \\
	\texttt{gianmarco.de.vita@usi.ch} \\
	\And
	\href{https://orcid.org/0000-0002-3037-8368}{\includegraphics[scale=0.06]{orcid.pdf}\hspace{1mm}Nargiz Humbatova} \\
	   Università della Svizzera italiana\\
            Lugano, Switzerland \\
	\texttt{nargiz.humbatova@usi.ch} \\
    \And
    \href{https://orcid.org/0000-0003-3088-0339}{\includegraphics[scale=0.06]{orcid.pdf}\hspace{1mm}Paolo Tonella} \\
	   Università della Svizzera italiana\\
            Lugano, Switzerland \\
	\texttt{paolo.tonella@usi.ch} \\
}

\begin{document}
\maketitle

\begin{abstract}
Testing Deep Learning (DL)-based systems is an open challenge. Although it is relatively easy to find inputs that cause a DL model to misbehave, the grouping of such inputs by features that make them problematic for the DL model under test is largely unexplored. In fact, existing approaches for DL testing introduce perturbations that may focus on specific failure-inducing features, while completely neglecting other ones that belong to different regions of the feature space.
%change?

In this work, we propose \topomap, a model-agnostic approach that relies solely on features that characterise inputs to construct a partitioning of the input space where each region represents a specific set of distinguishable input features.
%In this paper, we create an explicit topographical map of the input feature space, named \topomap, with the goal of exploring all regions that may potentially trigger a misbehaviour of a DL model. \topomap is black-box and model-agnostic, as it relies solely on features that characterise the input space. 
Such a partitioning gives users a possibility to identify and analyse areas that are under-represented, show higher failure rates, or exhibit prediction changes after a model update. The identified regions of interest can then be used for additional data generation or can help prioritise already generated inputs that require labelling efforts.

To discriminate inputs according to the specific features they share, we first apply dimensionality reduction to obtain input embeddings that are then subjected to clustering. Each DL model might require specific embedding computations and clustering algorithms, and specific configurations of the two, to achieve a meaningful separation of inputs into discriminative groups. We propose a novel way to evaluate alternative configurations of embedding and clustering techniques. We use a deep neural network (DNN) as an approximation of a human evaluator who could tell whether a pair of clusters can be discriminated based on the features of the included elements. We use such a DNN to automatically select the optimal topographical map of the inputs among all those that are produced by different embedding/clustering configurations.

The evaluation results show that the maps generated by \topomap consist of distinguishable and meaningful regions. In addition, we evaluate the effectiveness of \topomap using mutation analysis. In particular, we assess whether the clusters in our topographical map allow for an effective selection of mutation-killing inputs.
\end{abstract}


\section{Introduction}
As Deep Learning (DL)-based software solutions are spreading across a plethora of different application domains, testing of DL systems is progressively fostering related research \cite{Riccio2020} to ensure the integrity and reliability of those systems. In particular, one of the main goals is to identify the most effective test cases, i.e., inputs which bear features that are more likely to expose faults. %failures 

Existing approaches to DL input generation~\cite{7298640,8812047,10.1007/978-3-319-89960-2-22,7958570} often apply small input perturbations until a misbehaviour is exposed. However, these approaches may focus the test generation effort in a specific promising region of the input space, neglecting other, possibly less easily reachable regions, where different misbehaviours might be exposed. To the best of our knowledge, the only approach that explores the input feature space during test generation is DeepHyperion~\cite{10.1145/3460319.3464811,10.1145/3544792}. However, this approach requires substantial human effort for the definition and quantification of the feature space and requires a parametrised, model-based input generator that can be steered toward the exhaustive exploration of the input space.

The motivation for driving testing of DL software through the exploration of the input space, rather than using information extracted directly from the model under test, is twofold. 
First, the input space has a rich structure that can be used to meaningfully partition the regions of interest for testing. This can be a complementary, useful perspective w.r.t. input partitioning based on the inner model structure and its neuron activations.
Second, testing may not always involve a model the developer has access entirely to: this is the case of black-box and grey-box testing, when a third-party model is fed some inputs and produces an output, without revealing any of its internal structure/weights. There are  studies~\cite{10.1145/3460319.3464811,10.1145/3544792} leveraging the concept of \emph{illumination search}~\cite{mouret2015illuminatingsearchspacesmapping}, which characterise the input space as a grid and select, for each cell, the input maximising a  fitness function related to the model's faulty behaviour (e.g., distance from the decision boundary). %Such input-space-driven partitioning was shown to be quite useful and effective to expose model misbehaviours.

Such input-space-driven partitioning has demonstrated the connection between specific combinations of input features and their fault-exposing capabilities. % was fault-triggering
%\gianmarco{I think I am overciting DeepHyperion.}\gianmarco{Maybe I would remove any prior reference to input generation and move here the mention that DeepHyperion requires manual definition of features, something like: ``However, this approach requires substantial human effort for the definition and quantification of the feature space, needed to generate a map".}

\begin{figure*}[t]
    \centering
    \includegraphics[width=\textwidth]{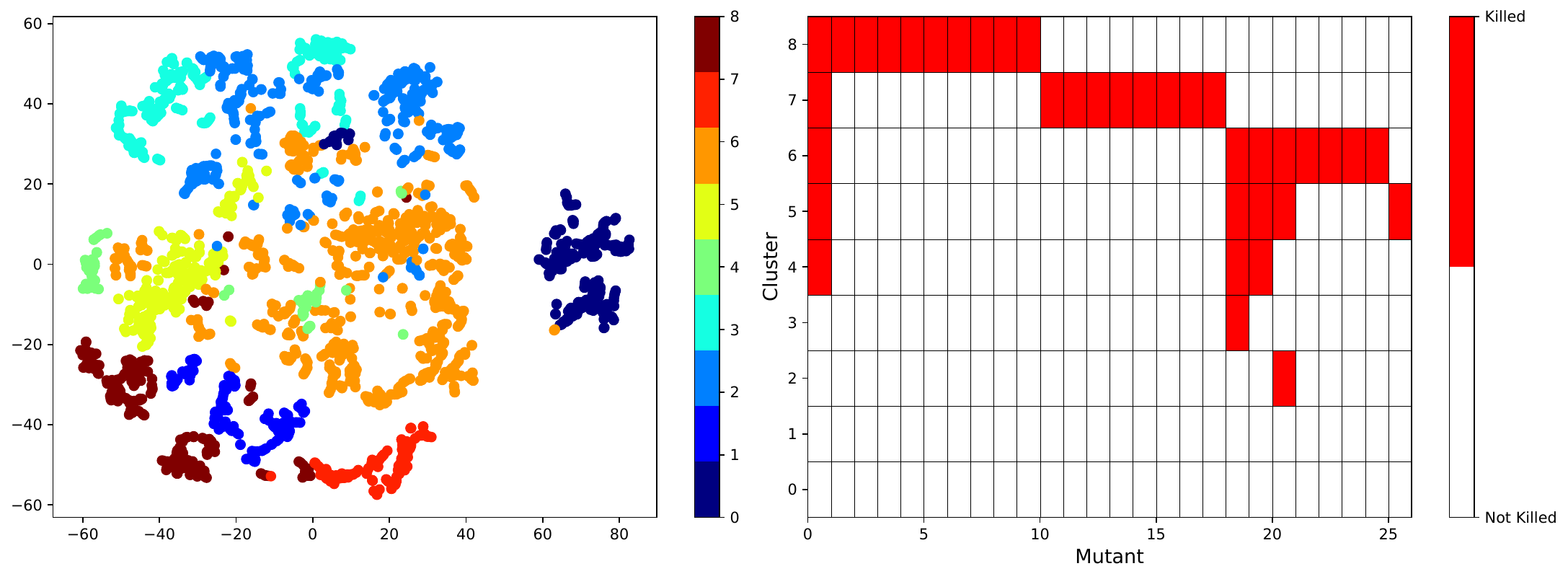}
    \caption{Topographical map of aggregations of test input clusters killing various mutants (\emph{Udacity} steering-angle prediction dataset)}
    \label{fig:demo_intro}
\end{figure*}

In this paper, we propose \topomap, a novel approach for fully automated construction of a topographical map of the input space, where inputs are grouped into regions according to their features. Starting from an initial available input pool, \topomap partitions the input space into distinct regions characterised by input features. This approach allows users to systematically explore input regions to identify combinations of input features that might be under-represented in the available dataset or are specifically hard for the model to learn (i.e., regions that exhibit higher failure rates) or analyse how the model behaviour changes on specific sets of inputs during the development process. Such observations could be useful to drive subsequent testing activities, for example, to generate new fault-revealing inputs or to prioritise large sets of unlabelled data. Moreover, once constructed, the \topomap acts as a persistent partitioning of the input space. As new test inputs become available, they are mapped onto the existing topology, allowing practitioners to continuously assess map coverage and failure distribution without rebuilding the map.

The input to a DL system is usually complex, in particular due to the high dimensionality of the data being processed~\cite{10.1162/neco-a-01490}. 
Hence, the first step of our approach is the automated extraction of an input embedding that has lower dimensionality than the input space, while still retaining the key features that characterise each input and trigger a specific behaviour of the DL system under test.

Second, we group the inputs according to the features represented in their embeddings using unsupervised learning methods, which infer the membership of the instances to a certain region (i.e., a cluster), based on the characteristics of the lower-dimensional data. 
As multiple embeddings and clustering algorithms/configurations are possible, we automate the selection of the topographical map by mimicking human assessment using a Deep Neural Network (DNN). A human may approach the quality assessment process for the output of a clustering by comparing pairs of clusters to decide if they contain discriminative features that make them recognisable and distinguishable groups of inputs. This task can be automated using a DNN, which is trained to classify the inputs according to the pseudo-labels (regions) produced by clustering. We assume that a high accuracy of this DNN correlates with the presence of meaningful, discriminative features in the two clusters compared. The output of this phase is a clustering of the inputs, represented as a topographical map (see Fig.~\ref{fig:demo_intro}, left).
It should be noticed that construction of the topographical map is model-agnostic, in that it is based uniquely on the analysis of the input domain, without considering any internal property of the DL model under test. Actually, the resulting topographical map is DL model-independent and can be reused across different model architectures operating on the same data.

The results indicate that the topographical maps constructed by \topomap contain regions that are clearly discernible and grounded in meaningful feature selection. Furthermore, the human study suggests that participants’ judgements are consistent with those of \topomap when separating inputs into map regions, particularly in cases where the dataset images are easy to interpret.

We empirically evaluated whether aggregations of \topomap's clusters are indicative of the presence of DL faults. %failures
To do this, we applied mutation analysis and aggregated the clusters based on their ability to \textit{kill} (i.e., expose the fault injected into) a mutated DL system. We define a failure as an observed misbehaviour of the DNN (e.g., an incorrect prediction on a given input), while fault refers to the underlying defect in the model responsible for such misbehaviour. In our mutation analysis, faults correspond to the artificial defects introduced by DeepCrime's mutation operators.

Fig.~\ref{fig:demo_intro} (right) shows an example in which aggregations made up of 1 to 5 clusters are capable of killing all mutants. The plot shows how the inputs responsible for triggering a misbehaviour in the tested DL system can be found to be concentrated in a few clusters, often just one. Finally, comparing side by side the two panels of Fig.~\ref{fig:demo_intro}, we can notice how the killing clusters exhibit some spatial proximity, e.g., in the case of the pair of clusters (5, 6), which supports our conjecture on the existence of a topographical structure behind the failure-inducing inputs. 

Our experimental results on a subject for which both tools are applicable show that the maps produced by \topomap differ substantially from those generated by the state-of-the-art approach DeepHyperion, revealing complementary structural representations of the input space. Moreover, \topomap enables effective partitioning of the input space, leading to the identification of regions denser in fault-revealing inputs, while requiring no manual feature engineering, highlighting its potential for scalable and automated testing of deep learning systems.
Besides that, our approach can identify mutation-killing inputs very efficiently achieving a killing probability of 100\% with just 9\% of the inputs, while random selection of the same number of inputs leads to a much lower killing (hence failure) probability of 65\%.
Compared to DeepHyperion, on the MNIST subject our approach achieves higher killing contributor density, exhibiting an 8\%\textit{pt} (resp. 12\%\textit{pt}) increase on killable (resp. initially non-killable)  mutants.

The paper is organised as follows: Sec.~\ref{sec:relatedwork} illustrates the recent advances in research on related fields. Sec.~\ref{sec:back} introduces some background on embedding computation and clustering. %\new{\sout{Sec.~\ref{sec:back} introduces some background on embedding computation and clustering.}}
Sec.~\ref{sec:approach} describes our approach, followed by its empirical assessment in Sec.~\ref{sec:exp}. Sec.~\ref{sec:results} analyses the results of the experiments, which are then discussed in Sec.~\ref{sec:discussion}, and finally, we draw our conclusions in Sec.~\ref{sec:concl}.
\section{Related Work}\label{sec:relatedwork}
In this section, we consider recent developments in research areas related to \topomap: model-agnostic testing of DNNs, input partitioning in traditional software testing,  prioritisation of the inputs exposing misbehaviours in a DL system, and  analysis of faults in DL systems. 
Finally, we compare our work with the most related one, DeepHyperion~\cite{10.1145/3460319.3464811,10.1145/3544792}.

\subsection{Model-agnostic Testing of DNNs}
Aghababaeyan et al.~\cite{aghababaeyan2023black} propose an approach to DNN testing that focuses on measuring diversity in test inputs. The authors selected 3 different diversity metrics for image-based datasets and evaluated their correlation with the fault detection capability of test inputs. The fault types are defined by applying a clustering technique to mispredicted inputs and assuming that each discovered cluster characterises a distinct DNN fault type. Their results show a positive and statistically significant correlation between geometric diversity metric~\cite{kulesza2012determinantal} and faulty regions in the DNN input space. The existence of such correlation supports our hypothesis on the effectiveness of input selection  based on specific combinations of input features. However, in our work we do not rely on diversity metrics, which are necessarily domain-specific (Aghababaeyan et al.~\cite{aghababaeyan2023black} investigated only image-based datasets). Our approach is more general, as it aims to separate test inputs into discriminative regions that emulate human-grade assessment without making any assumption on the existence of domain-specific diversity metrics. As a consequence, our experimental evaluation spans multiple domains, beyond image classification. 

\subsection{Input Prioritisation}
DL input prioritisation is a widely studied area, due to  the  high cost of labelling new inputs, which is often done manually and requires deep domain knowledge. Thus, when prioritising the inputs to test a DL system, it is important to automatically identify those that have high misbehaviour-revealing potential. 

Uncertainty quantifiers such as DeepGini~\cite{feng2020deepgini}, Vanilla Softmax~\cite{10.5555/3454287.3455541,hendrycks17baseline}, Prediction-Confidence Score (PCS)~\cite{zhang2020towards}, and Entropy~\cite{shannon1948mathematical}, estimate the uncertainty of a DL model based on the outputs of the \textit{softmax} layer, which makes these approaches only applicable to classification systems. Monte-Carlo Dropout~\cite{pmlr-v48-gal16}, however, can be used both for classification and regression tasks, as it uses the dropout layers of a model during the prediction phase to obtain multiple results for the same input and it evaluates uncertainty based on the discrepancies in the predicted values. Inputs with higher uncertainty are deemed as likely to be mispredicted.

Another family of approaches are based on surprise adequacy (SA)~\cite{kim2019guiding, 11026894}. These techniques focus on measuring how surprising an input is for a DNN, i.e., how different  the DNN’s activations produced on this input are, with respect to the activations observed for the training dataset. These techniques include Likelihood-based SA~\cite{kim2019guiding}, Distance-based SA~\cite{kim2019guiding}, and Mahalanobis-Distance Based SA~\cite{kim2020reducing}. The intuition behind the concept of SA is that the more surprising the input to the DNN, the higher its ability to trigger failures.

The last group of approaches relies on  neural coverage. Similar to traditional code coverage measures, metrics such as Neuron Coverage (NC)~\cite{pei2017deepxplore}, k-Multisection NC~\cite{ma2018deepgauge}, Neuron Boundary Coverage~\cite{ma2018deepgauge}, Strong Neuron Activation
Coverage~\cite{ma2018deepgauge} and Top-k NC~\cite{ma2018deepgauge} prioritise inputs based on their capability of triggering different DNN neuron activation patterns.

All these techniques require  access to the internal components of a DNN, such as the output of the softmax layer or neural activations. Our approach, however, is practically model-agnostic as it operates on the input features only. We create a discriminative topographical map of the inputs and we conjecture that different regions in such map can be aggregated to expose different DL faults. 

\subsection{Input Partition Testing}
In traditional software testing, among the test case selection strategies, an established and discussed family of techniques is \emph{partition testing} (PT)~\cite{10.1145/62959.62964}, which aims to subdivide the input domain into regions (subdomains) where inputs are assumed to show the same behaviour w.r.t. the software under test, in particular when it comes to fault exposure~\cite{10.1109/32.83906}. %PT approaches define subdomains of the input space according to some common properties that the inputs show w.r.t. the software under test~\cite{10.1109/32.83906}. %\nargiz{this sentence sounds a bit like a repetition of the previous one. Are there any more concrete examples or is it better just to drop it?}
The primary goal is to optimise the test suites by reducing the number of test cases to a set of representative cases from each partition, under the assumption of equivalence within partitions. PT strategies are regarded as an improvement over random input partitioning~\cite{10.1145/271775.271785}, as they introduce additional conditions to enhance testing effectiveness. For example, in \emph{proportional PT}, the number of selected inputs for each partition is proportional to the size of the corresponding subdomain~\cite{CHAN1996775}. %\nargiz{what is a subdomain? what is a probability of a subdomain? should we just leave size?}
More recent developments explore the use of reinforcement learning~\cite{Sun2025} to dynamically select test inputs based on feedback obtained from the software under test during execution.
%There are some works in the literature comparing PT strategies to random input partitions, leading to different conclusions~\cite{10.1145/271775.271785}. \nargiz{and what are the conclusions?} The sometimes negligible advantage of PT against the random partitioning \nargiz{do I understand it right that these techniques were never shown to be properly working?}\gianmarco{Most of the literature is old and I didn't manage to find more recent examples of it, but different works show that sometimes random partitioning still performs better than other techniques} \nargiz{the way it is written now reads rather strange. it creates an illusion the technique is useless. I would expand on 'different conclusions.}

None of the existing techniques have been applied directly on DL software. Furthermore, they are often based on criteria taking into account the program execution or its formal/informal specifications, while our approach generates a map relying exclusively on the input feature space in a totally model-agnostic fashion.

\subsection{DL Faults}
There is a number of studies that investigated faults in DL systems. In particular, Humbatova et al. performed an analysis of different sources  of DL faults (StackOverflow, GitHub, expert interviews) to produce a comprehensive taxonomy of real faults in DL systems~\cite{Humbatova:2020}. They define a DL fault as an inadequacy of a DL component to complete the task at hand that was caused by a human mistake during DL development or training. 
%The taxonomy was obtained by performing a manual analysis of 1059 artefacts collected from GitHub commits and issues, and from StackOverflow posts. The analysed artefacts were extracted from projects that use  widely adopted DL frameworks such as TensorFlow, Keras, and PyTorch. This source of information was complemented with semi-structured interviews with 20 DL practitioners with both academic and industrial backgrounds, during which the interviewees were asked about the types of faults they have experienced in their development practice. %The final taxonomy was validated through  a survey study with another set of 21 developers, which demonstrated that 87\% of the presented faults were encountered by at least 50\% of the respondents.

The works by Zhang et al.~\cite{Zhang:2018} and Islam et al.~\cite{Islam:2019} also mined faults from StackOverflow and GitHub platforms. Zhang et al.~\cite{Zhang:2018} focused on a set of applications that were developed using the TensorFlow platform and produced a set of 175 generic programming and DL-specific faults. Islam et al.~\cite{Islam:2019} considered a larger set of frameworks such as Theano, Caffe, Keras, TensorFlow, and PyTorch and collected $447$ different faults.

DeepCrime~\cite{10.1145/3460319.3464825} is a mutation testing tool designed for automated seeding of artificial faults (mutations) into DL systems. %Its main difference from DeepMutation++ is that DeepCrime is based on a set of mutation operators derived from \textit{real faults}. 
In DeepCrime the authors propose 35 and implement 24 \textit{source-level} mutation operators that target different aspects of the development and training of DL systems. This set of operators was extracted from an existing taxonomy of real faults in deep learning systems~\cite{Humbatova:2020} and was complemented with the issues found in the replication packages for the studies by Islam et al.~\cite{Islam:2019} and Zhang et al.~\cite{Zhang:2018}.

None of the existing works on the nature of DL faults attempted to characterise the input space regions that are capable of triggering a DL fault. The empirical evaluation of our approach relies on DeepCrime's mutants to draw such a connection.

\subsection{DeepHyperion}

DeepHyperion~\cite{10.1145/3460319.3464811,10.1145/3544792} is the most similar approach to \topomap available in the literature. DeepHyperion creates a feature map of the inputs based on a quantification of the structural and behavioural features that characterise the application domain. For instance, the structural features of the image of a digit may include its luminosity and orientation, while the behavioural features of a self-driving car may include its maximum distance from the centre of the lane. The definition of such features and their quantification require profound domain-specific expertise. The authors of DeepHyperion propose a methodology for the manual identification of relevant features and definition of metrics that quantify them. While we share with them the goal of finding discriminative features in the input space, our approach is fully automated. Indeed, one key design decision behind our approach is the usage of a DNN to mimic the human's discrimination capabilities. As a consequence, our approach has a wide range of applicability, which goes beyond the subjects considered in the evaluation of DeepHyperion.
\section{Background}\label{sec:back}

Our approach builds upon two well-established techniques: embedding computation and clustering. Below, we briefly summarise each to clarify their role in our pipeline.

\subsection{Embedding and Dimensionality Reduction}

In DL, many datasets contain high-dimensional data represented by hundreds to thousands of features. While highly effective for capturing complex and rich information, some of these features might be redundant or contain noise, while the high dimensionality of the data creates computational problems and hinders efficient downstream analysis and calculations. Embedding methods are important tools for transforming high-dimensional inputs, such as images or text, into more compact vector representations that capture meaningful semantic or structural information by mapping inputs from an original high-dimensional feature space to a lower-dimensional representation. These methods cover a wide range of techniques, spanning from linear dimensionality reduction to non-linear manifold-learning approaches.

Linear methods generate low-dimensional data representations by creating linear combinations of the original input features. In particular, PCA (Principal Component Analysis)~\cite{da6385d2-9c65-3860-bbcd-b821fdff69ff}, one of the most widely used linear dimensionality-reduction approaches, transforms the original features into a new set of orthogonal directions, called principal components, each of which is a linear combination of the original features. The principal components define a coordinate system oriented along directions of decreasing variance in the data. Initially, PCA can produce as many principal components as there are original features, while dimensionality reduction is achieved by retaining only the first $k$ components. The number of retained components is often chosen such that their cumulative explained variance exceeds a predefined threshold, for example 90\%~\cite{Jolliffe2002,HARROU2013129}. This reduces the dimensionality of the data while retaining most of its variability. Truncated Singular Value Decomposition (SVD)~\cite{Wall2003} is based on principles similar to those behind PCA. Specifically, SVD decomposes a given data matrix representing the dataset into a set of directions that capture patterns in the data. Each direction is assigned a singular value, which indicates how much variation in the dataset is captured by that direction. Dimensionality reduction is achieved by selecting the top $N$ directions with the largest singular values. Linear Discriminant Analysis (LDA)~\cite{Zhao2024} is another representative of linear methods that supports dimensionality reduction by incorporating information about class labels. Unlike PCA and SVD, when constructing directions, LDA favours separation between classes. Another approach, Independent Component Analysis (ICA), finds directions by maximising their statistical independence from each other. Finally, Random Projection (RP) uses randomly generated directions that preserve distances between data samples.

In some cases, linear transformations may not adequately preserve the relevant information in a dataset because the underlying structure of the data is characterised by non-linear relationships. An approach such as t-distributed Stochastic Neighbour Embedding (t-SNE) ~\cite{JMLR:v9:vandermaaten08a} constructs a lower-dimensional representation  that  preserves local neighbourhood relationships by minimising the difference between pairwise similarities of data points calculated in the original and embedded spaces. Because of this feature, it is successful in representing local clustering relationships and is often used for 2- and 3-dimensional visualisation. Uniform Manifold Approximation and Projection (UMAP)~\cite{10.1162/neco-a-01434} is another non-linear dimensionality-reduction technique that is capable of capturing complex relationships in data. Unlike t-SNE, it not only preserves local relationships but also retains some of the broader information about the structure of the whole dataset. This is achieved by connecting each data point to its nearest neighbours and identifying connectivity patterns across these local subsets. Isomap (ISO)~\cite{doi:10.1126/science.290.5500.2319} constructs a neighbourhood graph and estimates distances between samples as the shortest paths in this graph. A low-dimensional representation is constructed such that these distances are preserved as closely as possible.

Another class of embedding approaches relies on neural representations. One of the most widely used methods in this category is the Autoencoder (AE)~\cite{3045796.3045801}, which learns a compressed representation of the data by training a neural network to reconstruct its inputs and extracting the resulting latent-space representation. Variational Autoencoders (VAEs)~\cite{10.1561/2200000056}, in contrast, represent each input as a probability distribution in the latent space rather than as a fixed point.

There is also a class of neural embeddings designed for natural language processing. Word2Vec (W2V)~\cite{mikolov2013efficientestimationwordrepresentations} is one of the first approaches to construct static vector representations of words by considering the linguistic contexts in which words appear during the model training process. The core assumption is that words used in similar contexts have similar meanings. Thus, such words appear close to each other in the embedded space. FastText (FST)~\cite{bojanowski2016enriching}, which is similar to W2V, also considers parts of words (n-grams). GloVe (GLV)~\cite{pennington2014glove} analyses how often words occur together in the input text and constructs embeddings based on this information. BERT~\cite{BERT} learns representations of words that depend on the specific context in which they appear. By considering the surrounding words in a sentence, BERT can capture different meanings of the same word across different contexts. Consequently, unlike static embedding methods such as Word2Vec, the same word can be represented by different vectors depending on the context in which it is used.

%Other potential embedding procedures for the input data include extracting the activation vectors of the last hidden layers of the DNN under test (which would render the approach non-black box) or a feature extractor DNN. However, these approaches would introduce an additional layer of complexity, as the obtained embedding would depend not only on the data itself, but also on the proper training of the model from which the new reduced dimensions are extracted. Thus, we excluded them from our study.

In our approach, we explore the potential of a range of different embedding models in shaping topographical maps of input space.

\subsection{Clustering}

Clustering is a widely used unsupervised learning technique that groups data samples together into clusters based on the similarity of their features. The assumption is that samples that fall into the same cluster are more similar to each other than to samples forming other clusters. 

Flat clustering algorithms divide data samples into a given number of clusters ($k$). For example, for a given value of $k$, the K-means~\cite{https://doi.org/10.1348/000711005X48266} algorithm selects $k$ data points to represent the centres of the clusters (centroids) and populates the clusters by assigning input vectors to their closest centroid. Then, the algorithm computes new centroids by taking the average of the points within each cluster. It then iteratively recomputes the cluster membership of the data points to minimise variance within clusters until convergence.

Algorithms based on hierarchical clustering, in particular agglomerative approaches, begin the clustering process with each data sample representing a cluster, merging similar inputs together until they build a complete tree-like structure called a \emph{dendrogram}, where each inner node can be expanded into sub-clusters. The dendrogram can then be cut at a particular level to achieve the desired number of clusters. Divisive approaches, on the contrary, start with one cluster containing all samples and recursively divide it. The BIRCH clustering approach~\cite{10.1145/233269.233324} is another representative of hierarchical clustering; however, it is designed to work with larger datasets and is more efficient than other approaches in this family. It works by iteratively grouping similar samples together, starting with one data sample. As more samples are processed, they are compared to summary information (a clustering feature) representing each group instead of through pairwise comparisons with each individual data point. If a sample is too far from all existing groups, a new one is formed. The approach then builds a clustering feature tree, where clusters with similar features are placed together. Nearby clusters can be merged, or large clusters can be further divided as needed to reach the desired number of clusters if the latter is specified.

Density-based algorithms, such as HDBSCAN~\cite{10.1007/978-3-642-37456-2_14}, construct clusters by identifying regions in the input space that exhibit a high density of samples. Isolated samples coming from regions of low density are classified as noise. Unlike K-means, which assumes spherical cluster structures, density-based algorithms do not force the assignment of data points to specific clusters and allow the creation of clusters with irregular shapes. Such approaches do not allow direct specification of the desired number of clusters. This value can be indirectly manipulated by setting other algorithm parameters, such as the minimum cluster size or $min\_samples$, which determines how conservatively samples are classified as noise.

Spectral clustering builds a similarity graph that evaluates how similar the inputs in the dataset are. It then derives a similarity matrix from this graph and a degree matrix, a diagonal matrix that reflects the similarity degree for each sample. These two matrices are used to create a Laplacian of the similarity graph, and the algorithm proceeds by calculating its eigenvectors. These eigenvectors provide a representation of the samples that allows easier detection of the clustering structure.

The main idea behind exemplar-based clustering is to select specific data samples, called exemplars, and assign other samples to them. Similar to centroids in the K-means algorithm, exemplars are used to represent a group of inputs, with the difference that an exemplar is a real data point rather than an average of all inputs belonging to the group. Affinity propagation is a well-known exemplar-based clustering algorithm~\cite{doi:10.1126/science.1136800}. It starts by computing pairwise similarities between all data points, with each point considered a potential exemplar. The algorithm then iteratively measures how suitable each data point is as an exemplar and how much support a potential exemplar receives from other inputs. This process determines the final set of exemplars and clusters. Unlike K-means, this algorithm does not require the number of clusters to be specified in advance, as this value is determined internally.

Our approach adopts several clustering strategies to capture different forms of structure and organisation of the input space.
\begin{figure*}[!t]
    \centering
    \includegraphics[width=\textwidth]{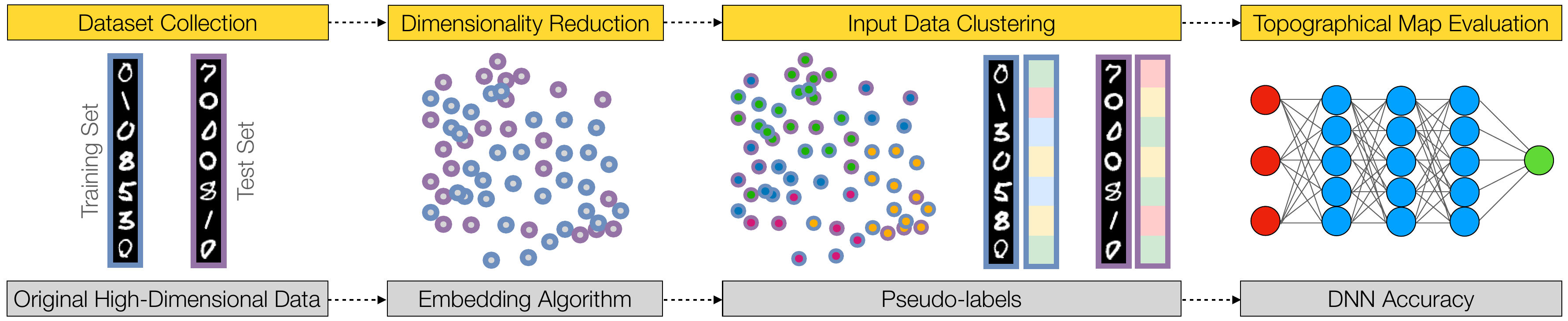}
    \caption{Cluster generation and evaluation pipeline}
    \label{fig:pipeline1}
\end{figure*}

\section{Approach}\label{sec:approach}

In this section, we describe \topomap, our approach for creating a topographical input map based on the input space features. The map can then be used to explore different regions of the input space and identify areas of interest, such as those that cause DL failures.%, to be used for the identification of regions that induce DL failures.

The main steps of \topomap are shown in Fig.~\ref{fig:pipeline1}. 
Given a set of inputs, split into training set and test set (Step 1: Dataset Collection), 
we compute a numerical embedding from the raw inputs (Step 2: Dimensionality Reduction). 
This is done for both the training set, which possibly includes a validation set (blue, in Fig.~\ref{fig:pipeline1}), and the test set (purple, in Fig.~\ref{fig:pipeline1}).
This step should ideally preserve the relevant features while reducing the input dimensionality, making it suitable for the next step — clustering. %Then, we apply a clustering algorithm, which, as a result, assigns each input vector a pseudo-label representing the cluster containing the input. 
More specifically, we take all instances of a given dataset as input, to
% Next, we merge the training set  and, if present, the validation set  keeping track of the original membership of each instance. Finally, we apply an embedding approach to the data. This 
ensure that the embedding is applied consistently to both the training (including the validation) and the test data.
Then, we apply a clustering algorithm that separates the inputs according to their features and assigns each input vector a pseudo-label representing its cluster (Step 3: Input Data Clustering). The clusters identified by the clustering algorithm outline distinct regions in the topographical map of the input space. After clustering the data, we divide it back into training (including validation) and test sets, according to the original split. In the final step (Step 4: Topographical Map Evaluation), the training (including validation) set is then used to train a DNN classifier that predicts the clusters' pseudo-labels. Finally, we use the trained DNN to automatically assess the quality of the map, by measuring its accuracy on the test set. 

It is important to note that the described pipeline contains multiple configurable parameters. The two most important ones are the choice of the embedding algorithm and the choice of the clustering algorithm. In addition, each specific embedding technique and clustering algorithm may have its own hyperparameters that the user has to select. We refer to each combination of embedding and clustering algorithm, including the specific selection of their hyperparameter values, as a \textit{clustering configuration}.

The aim of the topographical map quality evaluation step is to select the most cohesive map for a given input domain. Following the evaluation, the clustering configuration that produces the highest DNN prediction accuracy on the test set is reported to the user and the clusters it produces determine the resulting topographical map of the inputs. %through the capability of the DNN to assign a correct pseudo-label to the inputs.

Among the hyperparameters to be chosen for each clustering algorithm, the number of clusters $k$, which is required by some, but not all, clustering algorithms, deserves special treatment. In fact, running the entire pipeline depicted in Fig.~\ref{fig:pipeline1} for all possible values of $k$ is usually not affordable from a computational point of view, due to the typically large range of possible values of $k$. Hence, only for this hyperparameter we designed an ad hoc selection procedure, detailed in the next sub-section. Then, in the following sub-section we provide details about the last step of the pipeline in Fig.~\ref{fig:pipeline1}, automated assessment of the quality of the map associated with each considered clustering configuration (choice of $k$ excluded).

\subsection{Automated Selection of the Number of Clusters $k$}\label{sec:K_selection}
Some clustering algorithms, such as K-means, require the number of clusters, $k$, to be specified as an additional input. Finding an optimal number of clusters is crucial for our approach as this parameter directly influences the construction of the map and its correctness. However, it would be computationally too expensive to evaluate a large number of values for $k$ for use in our pipeline (see Fig.~\ref{fig:pipeline1}). To address this, we designed an ad hoc selection procedure for this specific hyperparameter only. The main idea is that clustering can be considered as a classification algorithm that operates by majority vote within each cluster, with the categorical labels (or numerical buckets, in case of a regression problem) of the original dataset serving as the ground truth to be predicted by the clusters. 
Specifically, each cluster is assigned a label determined as the majority label of the training set inputs in the cluster. Such a label is used as the cluster's prediction for all the test set inputs assigned to such a cluster. The predicted label is then compared to the ground-truth label, to determine the accuracy of the cluster's classification. 
We expect the clustering classifier to become more accurate as $k$ increases, because clusters become smaller and more cohesive. However, at some point, further subdivision of the data into smaller clusters becomes either unnecessary or only marginally beneficial. This happens because the common features that characterise each cluster have already been identified in previous steps (i.e., at lower $k$), and the new partitions are based on irrelevant and noisy features. Therefore, we can stop increasing $k$ when the plot of accuracy over $k$ flattens, as this indicates that the cluster classification capability has reached the saturation point. When this happens, we expect that all the features relevant for an optimal classification have been discovered in the current set of clusters. When dealing with regression problems, the ground-truth labels are represented by continuous numerical values rather than classes. Thus, to be able to apply the proposed automatic selection of $k$, we split the continuous values associated with the input data into buckets and treat each bucket as a class. We use the standard deviation of the continuous training %\nargiz{training? the distinction between train and test labels should be made clear here. i.e. where we use train and where we use test}
labels to determine the binning ranges, which define the boundaries of the buckets. %\paolo{explain how it is done or refer to an existing bucketing algorithm}

\begin{small}
\begin{algorithm}[t]
\caption{Selection of number of clusters $k$}
\label{algo:k_selection}
\small

\SetKwInOut{Input}{Input}
\SetKwInOut{Output}{Output}
\Input{$\mathcal{E}_\textit{train}$, $\mathcal{E}_\textit{test}$, embedded training and test set}
% \Input{$\mathcal{E}_\textit{valid}$, embedded validation set}
%\Input{$\mathcal{E}_\textit{test}$, embedded test set}
\Input{$y_\textit{train}$, $y_\textit{test}$, train and test set labels}
\Input{$n$, number of classes}
\Output{$k^*$, optimal number of clusters}

$\delta_0$ $\gets$ 0,
$\alpha_0$ $\gets$ 0,
$k_0$ $\gets$ $0$,
$k^*$ $\gets$ $n$\;

\While{True}{
    $m_{X}$ $\gets$ \textsc{ClusteringAlgorithm}($\mathcal{E}_\textit{train}$,\, $k^*$)\;
    $m_{C}$ $\gets$ \textsc{AssignClusterLabelsByMajority}($m_{X}$, $\mathcal{E}_\textit{train}$,\, $y_\textit{train}$)\;
    $y^C_\textit{test}$ $\gets$ $m_{C}(\mathcal{E}_\textit{test})$\;
    $\alpha_1$ $\gets$ \textsc{ComputeAccuracy}($y_\textit{test}$,\, $y^C_\textit{test}$)\;
    $\delta_1$ $\gets$ $\dfrac{\alpha_1 - \alpha_0}{k^*-k_0}$\;

    \If{$(\delta_1 + \delta_0) / 2 < 0.001$}{
        \textbf{break}\;
    }

    $\delta_0$ $\gets$ $\delta_1$,
    $\alpha_0$ $\gets$ $\alpha_1$,
    $k_0$ $\gets$ $k^*$,
    $k^*$ $\gets$ $k^* + n$\;
}

\Return  $k^*$\;
\end{algorithm}
\end{small}

Algorithm~\ref{algo:k_selection} presents the pseudo-code of our $k$ selection procedure.
We begin (Line 1) with an initial value of $k$ equal to the number of classes. Choosing a smaller value is not appropriate, as our goal is to further partition each class based on the distinguishing features exhibited by its inputs, rather than simply rediscovering the original class labels through clustering. Next, we perform clustering on the training set $\mathcal{E}_\textit{train}$ using the current value for the number of clusters $k^*$ (Line 3). 
Cluster labels are then assigned by majority vote to the unlabelled clusters of $m_{X}$ using the train set and its ground truth labels (Line 4). 
Since the original labels of the test set inputs ($y_\textit{test}$) are known, we treat the clustering output as a form of classification. We use the labelled clusters of $m_{C}$ to predict the labels of the test set inputs (Line 5) and compute the corresponding accuracy (Line 6), as well as its derivative (Line 7). If the derivative indicates that accuracy gains have plateaued (Line 8), the loop terminates and the current value of $k^*$ is returned. Specifically, we stop increasing $k$ when the average of the last two derivatives drops below 0.001, suggesting that the accuracy curve has likely flattened. If this condition is not met, $k^*$ is incremented by a multiple of the number of classes (Line 11), and the loop repeats (Line 2).It is important to note that the ground-truth labels are used exclusively for the automatic selection of the partition granularity ($k$). They play no role in other parts of the pipeline such as the selection of the embedding model or the assignment of inputs to clusters.

%\subsection{Creation of a topographical map} % Cluster generation and evaluation
There exist alternative approaches for the selection of the optimal number of clusters $k^*$, such as the Silhouette score~\cite{ROUSSEEUW198753}, the Davies-Bouldin index~\cite{4766909}, the Calinski–Harabasz index~\cite{doi:10.1080/03610927408827101} and the elbow rule~\cite{https://doi.org/10.1002,10.1145/3606274.3606278}.
However, these methods have well-documented limitations and rely on assumptions that may not hold for our datasets (e.g., convexity of the ground-truth clusters). Since we operate in a supervised deep learning context, where the model is tasked with predicting a label, we can leverage the available test set labels to assess the clustering quality for different values of $k$. Specifically, we evaluate how well each cluster groups vectors that share the same ground-truth label. In preliminary experiments, we explored the use of standard clustering quality metrics to guide the selection of $k$. However, none of the tested metrics consistently yielded reliable choices of $k$ across all datasets. This is likely due to violations of the underlying assumptions these metrics make, which are not satisfied by some of our datasets.
In contrast, the procedure described in Algorithm~\ref{algo:k_selection} consistently produced meaningful values of $k$ across all datasets, as confirmed by our manual inspection of representative clusters.

\subsection{Automated Assessment of the Quality of Clusters} \label{sec:qual}

As part of our approach, we propose an automated procedure to identify a pair of an embedding model and clustering algorithm (i.e., a \textit{configuration}) that produces the most cohesive and discriminative input space partition. We simulate the perspective of a human user who is asked to manually distinguish clusters based on the shared features of the elements within each group. %From this standpoint, we consider clusters to be of high quality if they are internally cohesive and clearly distinguishable from one another. 
From this standpoint, we consider clustering to be of a better quality if it defines regions that are internally cohesive and clearly distinguishable from one another. We use the performance of a DNN classifier as an automated proxy for the discriminability of the resulting clusters. While this proxy does not guarantee the best results from the semantic point of view, the resulting separability of the input regions was shown empirically to support several downstream testing tasks. %For example, a map whose regions are difficult to distinguish is less useful for estimating diversity of a pool of inputs by calculating the map coverage or for analysing localisation of failures. It is also harder to map new inputs to such a representation of the input space. 
Such a perspective can be viewed as an automated counterpart of human-subject evaluations, which have been adopted in related work focused on assessing the interpretability and consistency of input features in domains similar to ours~\cite{10.1145/3544792}.

For each clustering configuration, we apply the clustering model to the full dataset (both training and test set), and get a pseudo-label (i.e., a cluster identifier) for each input. This pseudo-label is assigned to the original, non-embedded data to serve as ground-truth label to be predicted by training a deep neural network (DNN) classifier. We  use the relabelled training (and possibly validation) set to train the DNN, and then we evaluate its performance on the corresponding relabelled test set. For each input domain, we choose a well-established and documented DNN architecture that proved to be effective in that particular domain, with adaptations applied as necessary to suit our specific classification task.

Since we cannot determine in advance whether the resulting clusters are balanced, and we actually expect them to be imbalanced in most cases, we apply class weighting during training. Specifically, we adopt a weighted training procedure to ensure that the loss function is not disproportionately influenced by the most populated classes. Furthermore, our focus is on evaluating the worst-case performance of the DNN, particularly its ability to distinguish between the pair of clusters that are most difficult to separate. In this context, the overall weighted accuracy may not provide sufficient insight, as it reflects an aggregate measure across all clusters rather than highlighting the most challenging distinctions. To address this, we introduce a novel metric, referred to as weighted pairwise accuracy, which effectively projects standard accuracy onto each possible pair of clusters. We then consider the minimum value across all pairs as a representative indicator of the DNN’s worst-case performance.

We evaluated the DNN’s ability to distinguish between each pair of clusters using the \textit{weighted pairwise accuracy metric}. 
For a pair $\langle A, B\rangle$, we compute its weighted pairwise accuracy as:

\begin{equation} \label{eq:pwa}
w_\textit{acc}^{\textit{A,B}} =
\begin{cases}
\dfrac{w\cdot n_\textit{AA} + n_\textit{BB}}{w\cdot n_\textit{AA} + n_\textit{BB} + w\cdot n_\textit{BA} + n_\textit{AB}}, \quad w = \dfrac{|B|}{|A|}, \quad \text{if } |A| < |B|\\[10pt]
\dfrac{n_\textit{AA} + w\cdot n_\textit{BB}}{n_\textit{AA} + w\cdot n_\textit{BB} + n_\textit{BA} + w \cdot n_\textit{AB}}, \quad w = \dfrac{|A|}{|B|}, \quad \text{otherwise.}
\end{cases}
\end{equation}
    
\noindent
where $n_\textit{AA}$ (resp. $n_\textit{BB}$) indicates the total number of instances belonging to cluster $A$ (resp. $B$) that are correctly classified, while $n_\textit{BA}$ (resp. $n_\textit{AB}$) indicates the number of elements of cluster $A$ (resp. $B$) being predicted as instances belonging to cluster $B$ (resp. $A$). As this computation is conducted pairwise, any further instance of the two clusters that are labelled neither $A$ nor $B$ is discarded when considering the pair $\langle A, B\rangle$. %\nargiz{the meaning of the last sentence is not clear to me}
For the purpose of computing $w_\textit{acc}^{\textit{A,B}}$, the size of cluster $A$  (resp. $B$) is determined as $|A| = n_\textit{AA} + n_\textit{BA}$ (resp. $|B| = n_\textit{BB} + n_\textit{AB}$). It can be noticed that the definition of weighted accuracy $w_\textit{acc}^{\textit{A,B}}$ is symmetric w.r.t. the pair $\langle A, B\rangle$, i.e., $w_\textit{acc}^{\textit{A,B}} = w_\textit{acc}^{\textit{B,A}}$. Consequently, we only need to compute it for half of the pairs $\langle A, B\rangle$ with $A \neq B$, and the remaining pairs can be obtained by symmetry.

\begin{figure*}[t]
    \centering
    \includegraphics[width=\textwidth]{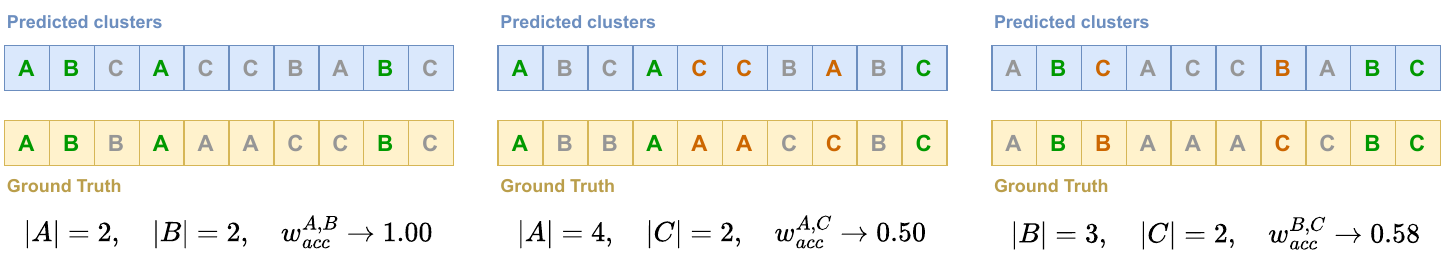}
    \caption{Computation of the \emph{weighted pairwise accuracy} on a three-cluster prediction example}
    \label{fig:pwacc}
\end{figure*}

% For consistency with the introduction Figure -> Fig.
Fig.~\ref{fig:pwacc} shows an example of computation of weighted accuracy for all pairs of 3 clusters, namely $\langle A, B\rangle$, $\langle A, C\rangle$, $\langle B, C\rangle$. For the first pair, $\langle A, B\rangle$, 4 entries (in green) are taken into account, i.e., only those that contain either of the two clusters in both the ground truth and the prediction.  
It should be noticed that we discard entries that contain cluster $B$ in the ground truth, but $C$ in the prediction (i.e., the 3rd entry) or $A$ in the prediction, but $C$ in the ground truth (i.e., the 8th entry). While these entries do not contribute to the weighted accuracy for the pair $\langle A, B\rangle$, they will contribute to the other pairs (specifically, the 3rd entry to $\langle B, C\rangle$, the 8th to $\langle A, C\rangle$). So, they are taken into account later.
On the remaining pairs, we measure the DNN weighted accuracy, which is 100\% as all predictions are correct (all 4 kept entries are green). 
In this case, weighted accuracy and normal accuracy coincide, as $|A|$ is the same as $|B|$. When considering the pair $\langle A, C\rangle$, 6 entries are kept (3 correct/green; 3 incorrect/red), while 5 entries are kept for the pair $\langle B, C\rangle$ (3 correct/green; 2 incorrect/red). In both cases, $|A|$ and $|B|$ differ, so we need to adjust for the class imbalance by introducing a weight $w$, equal to 2 in the first case and to 3/2 in the second. This weight multiplies the number of ground truth entries of the minority class, to account for the imbalance. Hence, for $\langle A, C\rangle$, $w_\textit{acc}^{\textit{A,C}} = (2 + 2 \cdot 1) / (2 + 2 \cdot 1 + 2 + 2 \cdot 1) = 1/2$. For $\langle B, C\rangle$, $w_\textit{acc}^{\textit{B,C}} = (2 + 3/2 \cdot 1) / (2 + 3/2 \cdot 1 + 1 + 3/2 \cdot 1) = 7/12$. 

The minimum of such pairwise accuracies, i.e., 0.5 for the pair $\langle A, C\rangle$, is an indicator of the worst-case DNN's ability to identify discriminative features in the clusters. We use such minimum pairwise accuracy as selection criterion for choosing the best clustering configuration.

\begin{small}
\begin{algorithm}[t]
  \caption{Selection of clustering configuration}\label{algo:tm}
  \small
  \SetCommentSty{mycommfont}
    \SetKwInOut{Input}{Input}
    \SetKwInOut{Output}{Output}
    \Input{$CC = \{\langle E_1, C_1\rangle, \langle E_2, C_2\rangle, \ldots\}$, candidate clustering configurations}
    \Input{$X_\textit{train}$, $X_\textit{valid}$, $X_\textit{test}$, original training, validation and test set}
    %\Input{$\langle E, C, k^* \rangle$, configuration of embedding, clustering algorithm and number of clusters}
    %\Output{$w_{\textit{acc}}^{A,B}$, $\min w_{\textit{acc}}^{A,B}$, average and minimum pairwise accuracy}
    \Output{$\langle E^*, C^*\rangle$, optimal clustering configuration}
    
  \SetKwProg{Def}{def}{:}{}
  %\Def{\textsc{SelectTopographicalMap}($X_\textit{train}$, $X_\textit{valid}$, $X_\textit{test}$)}{
    $\textit{Min} \gets \emptyset$\;
    \ForEach{$\langle E, C\rangle \in CC$}{
        $m_{C}$ $\gets$ $C$($E$($X_\textit{train}$))\;
        $y^C_\textit{train}$, $y^C_\textit{valid}$, $y^C_\textit{test}$ $\gets$ $m_{C}$($E$($X_\textit{train}$)), $m_{C}$($E$($X_\textit{valid}$)), $m_{C}$($E$($X_\textit{test}$))\;
        $m_\textit{DNN}$ $\gets$ \textsc{TrainDNN}($X_\textit{train}$, $y^C_\textit{train}$, $X_\textit{valid}$, $y^C_\textit{valid}$)\;
        $\textit{pred}^\textit{C}_\textit{test}$ $\gets$ {$m_\textit{DNN}$}($X_\textit{test}$)\;
        $\textit{min\_acc} \gets 1$\;
        \ForEach{$\langle A, B\rangle \in m_{C}$.clusterIds $\times m_{C}$.clusterIds, $A \neq B$}{
            %$\langle$ $\langle E, C, k^* \rangle$, $w_{\textit{acc}}^{A,B}$, $\min w_{\textit{acc}}^{A,B}$ $\rangle$ $\gets$ \textsc{ComputePairwiseAccuracy}($\textit{pred}^\textit{C}_\textit{test}$, $y^C_\textit{test}$)\;
            $\textit{acc}$ $\gets$ \textsc{ComputePairwiseAccuracy}($\textit{pred}^\textit{C}_\textit{test}$, $y^C_\textit{test}$, $A$, $B$)\;
            \lIf{$\textit{acc} < \textit{min\_acc}$}{$\textit{min\_acc} \gets \textit{acc}$}
        }
        $\textit{Min} \gets \textit{Min} \cup \{\langle E, C, \textit{min\_acc}\rangle\}$\;
    }
    \Return{$\langle E^*, C^*\rangle$ such that $\langle E^*, C^*, \textit{min\_acc}\rangle \in \textit{Min}$ and   $\textit{min\_acc} = \max \{\textit{Min}.\textit{min\_acc}\}$}\;
    %}
\end{algorithm}
\end{small}

Algorithm~\ref{algo:tm} shows the pseudo-code of our procedure for the selection of the most discriminative clustering configuration. The input $CC$ is a set of clustering configurations $\langle E_i, C_i\rangle$, where $E_i$ represents an embedding and $C_i$ a clustering algorithm, both instantiated with a concrete choice of hyper-parameters. Algorithm~\ref{algo:tm} takes also in input the original (pre-embedding) input vectors, split into training, validation and test set. 

We use set $\textit{Min}$, initialized at Line 1, to collect the minimum weighted pairwise accuracy of each configuration. For each configuration $\langle E, C\rangle$ (Line 2), we apply the clustering algorithm $C$ to the training set transformed through the embedding $E$ (Line 3). We use the resulting clustering model $m_C$ to assign a pseudo-label to the input vectors in training, validation, and test set (Line 4) and we train a neural network $m_\textit{DNN}$ using training and validation set (Line 5). We predict the labels of the test set vectors at Line 6, and then we iterate over all pairs of clusters to determine the pair with minimum weighted accuracy (loop at Line 8). In particular, at Line 9 we compute the pairwise weighted accuracy for each pair $\langle A, B\rangle$ and at Line 12 we assign the minimum value to the current configuration $\langle E, C\rangle$, by storing the triple $\langle E, C, \textit{min\_acc}\rangle$ into set $\textit{Min}$. Finally, we return as optimal configuration $\langle E^*, C^*\rangle$ the one that maximizes the worst case pairwise accuracy (Line 14).

\section{Experimental Procedure}\label{sec:exp}

This section describes the evaluation procedure used to assess the topographical maps produced by \topomap, including the research questions that form the basis of our empirical assessment.

\subsection{Subjects}

In order to attain a comprehensive evaluation of our approach, we considered datasets pertaining to different domains, representative of different types of DL applications, spanning from image classification to the prediction of angles through regression. As currently there is no comprehensive and realistic benchmark of DL faults available, we adopt faulty models generated by the DeepCrime mutation tool and available in its replication package~\cite{10.1145/3460319.3464825}. 
We consider only killable mutants, as their performance can be distinguished from that of the original model by a set of test inputs.

Four of the datasets considered solve classification problems: (i) the \emph{MNIST} (MN) dataset~\cite{data:mnist}, consisting of $28\times 28$ greyscale images of ten handwritten digits. The digit classifier is implemented as an eight-layered convolutional neural network (CNN)~\cite{model:mnist}; (ii) the \emph{CIFAR-10} (CF) dataset~\cite{data:cifar10}, consisting of $32\times 32$ colour images belonging to ten different classes. The used DL system is a 2D CNN; (iii) the Kaggle \emph{Speaker Recognition} (SR) dataset~\cite{data:speakerrecognition}, consisting of audio speech samples from five different politicians, labelled with their respective names. The classification task is carried out by deriving a frequency domain representation of the input samples through the Fast Fourier Transform and then training a 1D CNN model~\cite{model:speakerrecognition}; (iv) the \emph{Reuters} (RE) dataset from Keras~\cite{chollet2015keras}---a preprocessed version of the original data collected by Lewis~\cite{miscreuters}---consisting of a selection of news articles labelled by topic. Each entry is modelled as a high-dimensional binary vector marking with a 1 the words present in the document (one-hot vocabulary encoding). 

The other two datasets pertain to regression problems. In these two cases, only for the automated selection of the optimal number of clusters $k^*$%\changed{(for the mutation-based evaluation the continuous labels are kept)} %Too defensive imho
, we map each numeric output to a set of discrete buckets, determined based on the %mean $\mu_y$ and 
standard deviation $\sigma_y$ of the output $y$ across the entire dataset. Specifically, for the bucketing of each output $y$ we use the following three ranges: %$(-\infty, \mu_y - \sigma_y), [\mu_y - \sigma_y, \mu_y + \sigma_y], (\mu_y + \sigma_y, +\infty)$. 
$(-\infty, - \sigma_y), [- \sigma_y, + \sigma_y], (+\sigma_y, +\infty)$. In the case of UE, where the output labels $y_1, y_2$ belong to $\mathcal{R}^2$, we apply these three ranges to each dimension, thus obtaining nine distinct buckets.

%For UE instead, the inputs are disposed into a 2D data structure, for which we set up nine buckets basing upon the following nine ranges: $\langle (-\infty, - \sigma_y),    (-\infty, - \sigma_y) \rangle$, $\langle (-\infty, - \sigma_y),    [- \sigma_y, + \sigma_y]\rangle$, $\langle (-\infty, - \sigma_y),    (+\sigma_y, +\infty)\rangle$,  $\langle [- \sigma_y, + \sigma_y], (-\infty, - \sigma_y) \rangle$, $\langle [- \sigma_y, + \sigma_y], [- \sigma_y, + \sigma_y]\rangle$, $\langle [- \sigma_y, + \sigma_y], (+\sigma_y, +\infty) \rangle$, $\langle (+\sigma_y, +\infty), (-\infty, - \sigma_y) \rangle$, $\langle (+\sigma_y, +\infty), [- \sigma_y, + \sigma_y]\rangle$, $\langle (+\sigma_y, +\infty), (+\sigma_y, +\infty) \rangle$,

\begin{comment}
SOURCE CODE UD: 
    std_array = np.std(y_train)

    bucket1 = []
    bucket2 = []
    bucket3 = []

    for ind, element in enumerate(y_train):
        if element < -std_array:
            bucket1.append(element)
        elif -std_array <= element <= std_array:
            bucket2.append(element)
        elif element > std_array:
            bucket3.append(element)
\end{comment}

The regression problems considered are: (v) the \emph{Udacity} (UD) dataset~\cite{9438556}, consisting of images of urban roads labelled with the steering angle used by a self-driving car model to learn how to keep the car in the lane. The DL model we employ is the one implemented by NVIDIA~\cite{bojarski2016end}; (vi) the \emph{UnityEyes} (UE) dataset~\cite{model:unity}, consisting of synthetic eye images rendered by the Namesake framework~\cite{10.1145/2857491.2857492}, with the target learning model, implemented as a CNN based on the LeNet architecture~\cite{model:unity}, being an estimator of the 2D eye gaze direction angle. For what concerns the assessment of the correctness of a prediction, required to determine the contributing inputs for the UD dataset, we adopt the threshold $\tau = 0.3$, following previous empirical work~\cite{10.1145/3460319.3464825}. For UE we instead set a threshold $\tau = 5^{\circ}$ which is the step for both yaw and pitch angles used during the generation of the dataset.

\subsection{Research Questions}
We have performed a set of experiments to answer the following research questions:

%\textbf{RQ1 (Instances Discrimination):} \emph{Does the proposed approach generate clusters grouping together instances with identifiable and discriminative features?}
\textbf{RQ1 (Discrimination):} \emph{How useful is the proposed approach to select the optimal clustering configuration, i.e., the clustering configuration that better discriminates the test input instances?}

The goal of this research question is to determine whether different clustering configurations achieve different degrees of cluster separability and cohesiveness, and whether the same configurations perform best across different subject systems.
To identify the clustering configuration that produces the most cohesive and discriminative clusters, we evaluate combinations of embedding models, clustering algorithms, and their corresponding hyperparameters.

\textbf{Experimental Design.} To achieve the above-stated goal we use the DNN-based cluster evaluation pipeline (see Fig.~\ref{fig:pipeline1} and Sec.~\ref{sec:qual}) for the selection of the optimal clustering configuration under the assumptions that (1) not all clustering configurations are equivalent; and (2) the optimal clustering configuration might be different across different DL problems (regression vs classification)/domains/subjects.

\textbf{Metrics.} We use the minimum weighted pairwise accuracy (see Eq.~\eqref{eq:pwa}) to characterise the ability of each clustering configuration to produce clusters that a DNN classifier can discriminate. For the sake of completeness, we also include several well-established clustering quality metrics, such as the Silhouette score~\cite{ROUSSEEUW198753}, the Davies-Bouldin index~\cite{4766909}, the Calinski–Harabasz index~\cite{doi:10.1080/03610927408827101}.

\textbf{RQ2 (Human Alignment):} \emph{How does the cohesion and separability evaluation of the DNN assessor align with that of humans?}

Our intuition behind a DNN selecting the best topographical configuration lies in the idea that it may simulate a human  assessing whether two clusters are distinguishable or not. 
In this RQ, we want to determine whether human assessors are able to identify patterns that differentiate \topomap's clusters in a way that correlates with the automated assessment made by \topomap's DNN assessor. For this purpose, we conducted a human study to assess whether the DNN classifier can provide a reliable approximation of the human's ability to recognise and distinguish clusters. We intentionally base our evaluation on a sample of cluster pairs with different pairwise accuracies, ranging from pairs that are easy for the DNN to distinguish to those that are more difficult.

We chose MN and CF as subjects as this study requires datasets for which humans can reasonably recognise and discriminate the underlying input features. We therefore selected the two image-based datasets used in this work to cover different levels of visual complexity. In particular, MNIST represents a relatively simple visual domain, whereas CIFAR-10 contains more complex natural images.

The remaining subjects considered in our evaluation are not well suited to the adopted experimental procedure. Reuters is natural text, while the Speaker Recognition dataset would require an audio-based study. The Udacity and UnityEyes datasets also pose challenges for human assessment. In UnityEyes, each sample consists of an eye image associated with a head/camera rotation angle, making it difficult for participants to identify the underlying characteristics that might define different clusters. Similarly, the Udacity dataset comprises highly compressed driving simulator frames in which most of the image depicts the road ahead, and the visual differences between inputs are often subtle and difficult for humans to recognise and consistently group.
Thus, we restricted the human experiment to MN and CF, as these subjects provide a higher likelihood of obtaining meaningful, non-random responses from participants.
%We constructed four questionnaires, two for each of the selected subjects. In total, we collected 20 responses for each questionnaire. 

\textbf{Experimental Design.}
%\subsection{\new{Evaluation: Human Assessment}} %\label{sec:human-study}
%The primary objective of the pipeline that generates a topographical map is to identify input space regions that are both recognisable and distinguishable. To evaluate this, we investigate whether the pipeline focusses on features that are interpretable by humans. For example, in the domain of digit classification, such features might include the orientation or boldness of the digits. To this end, 
We perform a user study to understand whether manual human assessment agrees with the clustering produced by our approach. When assigned the task of separating distinguishable inputs into two groups, human assessors would ideally produce a clustering consistent with that of \topomap. On the other hand, when the input is hard to distinguish, both a human and \topomap are expected to struggle to achieve high separation accuracy. 

\begin{figure*}[t]
    \centering
    \includegraphics[width=\linewidth]{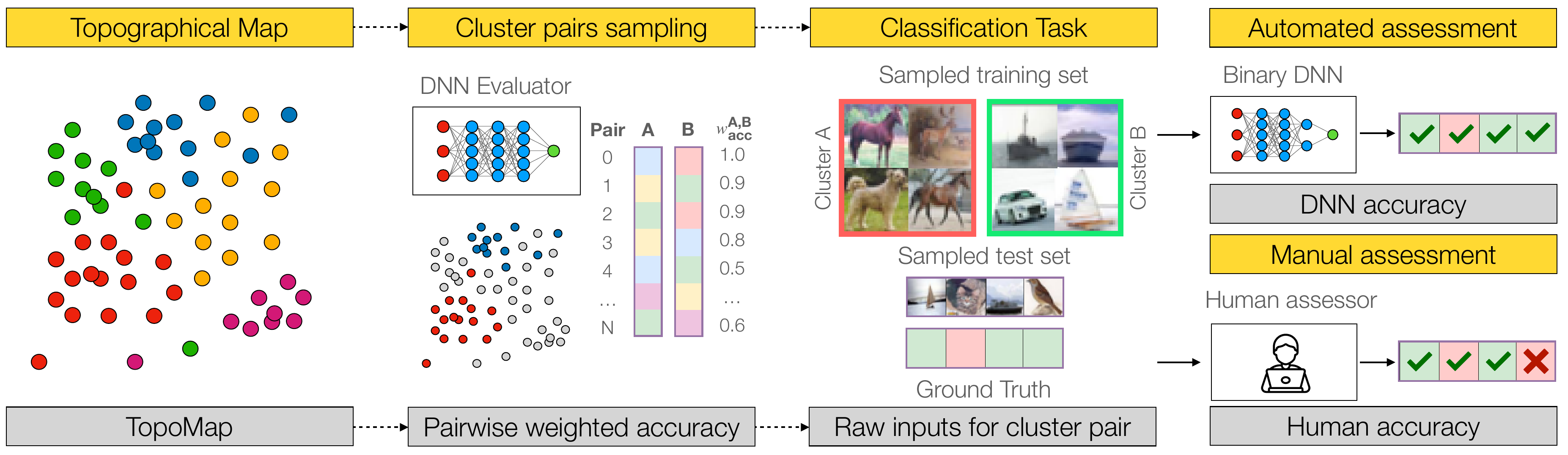}
    \caption{Empirical design of the human evaluation study}
    \label{fig:empirical_human_study}
\end{figure*}
 
 The idea behind the design of the human study is presented in Fig.~\ref{fig:empirical_human_study}. Given a topographical map, we identify the pairs of clusters and rank them by $w_\textit{acc}^{\textit{A,B}}$. To allow a thorough evaluation, we use a selection of inputs from pairs of clusters with different values of $w_\textit{acc}^{\textit{A,B}}$. In other words, we carefully select diverse pairs of clusters that are easy, moderate, or difficult to distinguish. For each cluster in a pair, we sample a fixed number (15) of training set inputs and present them to human assessors to give them an idea of how clusters were formed. Then, for each pair, we pick 10 test set inputs from each cluster and ask a human assessor to assign each input to one of the two clusters. We then compare the accuracy of the human assessor with that of the DNN used by \topomap. Furthermore, we also compare it with the accuracy of a binary DNN trained only on the sampled cluster pair, which represents an empirical upper bound for accuracy. We also ask the human assessor for each pair to indicate how difficult the task was, on a scale from 1 to 5, with the difficulty being explained as the arbitrariness of the choice when assigning an image to a cluster (the more the choices are arbitrary, the more the classification task is difficult). Fig.~\ref{fig:empirical_human_study} illustrates the procedure. 
 In the first step, we construct a topographical map of the inputs using the \topomap approach. %described in the previous section. 
 In Step 2, we consider all possible pairs $A, B$ of clusters and rank them by the \topomap's weighted accuracy, obtained from the DNN evaluator of \topomap. Among them, we take a subset so as to cover pairs that are easy, moderate and difficult to separate according to the DNN evaluator of \topomap. In practice, we sample these pairs from the initial, middle and final parts of the list ranked by weighted accuracy.
 In Step 3, for each sampled pair of clusters $A$ and $B$ (resp. red and green), we create a classification task, consisting of examples of inputs selected from the training set and belonging to both clusters, along with a set of inputs from the test set belonging to those clusters and to be classified automatically or manually. 
 The ground truth for the sampled test inputs is the cluster (by construction, $A$ or $B$) they belong to.
 Then, in Step 4, both the DNN and the human assessor classify the test inputs into clusters A or B. These classifications are then checked against the ground-truth labels, enabling a comparison of the two accuracies.

 %In the case of MNIST, the $w_\textit{acc}^{\textit{A,B}}$ of the 30 chosen sample pairs ranges from 0.77 to 1.00, while for CIFAR-10, it ranges from 0.58 to 1.00. In both cases, the sample distributions are generally mesokurtic, thus similar to a normal distribution. 
The study was structured in the form of a questionnaire consisting of 15 distinct tasks. %where at each task, a pair of distinct clusters is shown and the user has to determine which group a selection of inputs drawn from the test set belongs to (the inputs belong to either one of the two clusters).
In addition, we included an attention check to ensure that responses are not given at random. The attention check is not disclosed to the human assessor and consists of a single trivial task, such as separating multiple copies of the same handwritten bold sample of digit `2' from narrow and tilted samples of digit `0'. To guarantee a range of diverse and significant responses, a total of four questionnaires have been designed, two based on handwritten digit recognition (MN) and the other two on image classification (CF). We designed our questionnaires using Qualtrics~\cite{qualtrics}, an online platform commonly used to conduct surveys, evaluations, and other forms of data collection. Participants were recruited through the Amazon Mechanical Turk~\cite{mturk} platform. Amazon Mechanical Turk allows users to filter participants by the level of performance measured on a previously performed task. We chose an approval rate threshold of 95\% in accordance with previous studies~\cite{Peer2014}. Our aim was to recruit 20 participants for each questionnaire. After collecting the responses, we kept only those that passed the attention check and republished questionnaires until the target number of valid responses was obtained.

\textbf{Metrics.} To evaluate the correlation, we compute the Pearson correlation coefficient (PCC) and the associated $p$-value between the human accuracy ($h_{acc}^{A,B}$) and (i) the weighted pairwise accuracy of \topomap's DNN assessor ($w_{acc}^{A,B}$), (ii) the accuracy of a DNN trained and tested only on the pair of clusters under consideration ($t_{acc}^{A,B}$), (iii) the accuracy of the latter DNN on the same exact 10-input sample analysed by the user (${t'}_{acc}^{A,B}$). We compute the same correlation statistic also for the human perceived difficulty ($h_{\textit{diff}}^{A,B}$).\\

\textbf{RQ3 (Fault Detection):} \emph{Are clusters indicative of faults in DL systems?}

Our goal is to determine whether the clusters identified by \topomap contain inputs that can effectively discriminate between faulty (in our experiments, mutant) and correct (in our experiments, original) DNN models, i.e., whether \topomap identifies fault-revealing features. 

In our experiments, we use mutants as proxies for faulty DL models. Specifically, we consider mutants generated by DeepCrime~\cite{10.1145/3460319.3464825}, a pre-training DL mutation tool based on real DL faults. DeepCrime implements 24 mutation operators (MOs), each of which can be configured by setting appropriate parameter values. For example, the mutation operator ``change learning rate'' can be configured by specifying a mutated learning rate value.

The set of applicable DeepCrime mutation operators depends on the architecture and training procedure of the subject model. Several DeepCrime operators target architectural components that are absent from some of our subjects. For example,  dropout-specific operators cannot be applied to models without dropout layers. Likewise, the RCP operator is not applicable to subjects that do not use an early-stopping callback. Consequently, not every mutation operator is applicable to every subject. For each subject, we therefore considered all DeepCrime mutation operators that were applicable to the corresponding model architecture, following the original DeepCrime methodology. As a result, out of 24 available mutation operators, 17 were applicable to our subjects:
%In this study, we focus on the following mutation operators:
add activation function to layer (AAL), add noise to training data (TAN), add weights regularisation (RAW), change activation function (ACH), change batch size (HBS), change early-stopping patience (RCP), change number of epochs (HNE), change labels of training data (TCL), change learning rate (HLR), change loss function (LCH), change optimisation function (OCH), change weights initialisation (CWI), remove a portion of training data (TRD), disable data batching (HDB), make output classes overlap (TCO), remove activation function (ARM), and unbalance training data (TUD).  

The mutants generated with DeepCrime differ both in their area of application and in the effect they have on the model (e.g., on accuracy). Some mutations can severely affect the model's ability to generalise, leading to a uniform degradation across the input space, while some introduce only subtle changes. For example, besides targeted mutations affecting specific subsets of the training data, mutations affecting how the training procedure unfolds (e.g., changing weight initialisation) may still induce localised failures. This happens because certain input regions contain examples that are intrinsically more difficult to learn as they exhibit more subtle or ambiguous feature combinations, while others are easier to classify. A globally altered training process may therefore disproportionately affect these more challenging regions, resulting in failures concentrated within specific parts of the input space. For each dataset, we evaluate the ability of test set instances to expose mutants generated from the original high-performance DL model. In our evaluation, we restrict the analysis to mutants that are killable by the entire test set. %The structure of the evaluation pipeline is shown in Fig.~\ref{fig:pipeline2}.

%This question aims to assess whether the clusters obtained from \topomap group together instances that share features  making them capable of triggering misbehaviours in the tested DL systems. In particular, we are interested in determining whether there exist small cluster aggregations that can kill each mutant of the original DL system thanks to a high density of killing contributors. 
%\nargiz{paraphrase a bit}

We deem a mutant killing cluster aggregation as \textit{fault-discriminative} %failure-discriminative
if only a small number  of clusters is sufficient to create it and if it contains a high density of mutant killing contributors. 
In fact, the presence of only a few contributors in a cluster may not necessarily result in statistically killing the mutant, as their influence may be diluted by the  other non-contributing inputs, which means a larger and denser cluster aggregation should be formed to possibly kill the mutant.

We compare the killing capability of each cluster aggregation with a random selection of a set of inputs having the same size as the aggregation. The cluster aggregation is regarded as a high density failure region if the corresponding random selection has a substantially lower probability of killing each mutant. In fact, this would indicate that developers choosing the test inputs based on our topographical map have a higher chance of exposing faults %failures
of the DL system than developers who choose the test inputs randomly. 

We conduct this experiment on a set of killable mutants, namely mutants that can be killed by the available test data. Additionally, we checked whether our approach can locate a region on the map that successfully kills mutants that cannot be killed by the full test set, i.e., when the number of contributing inputs in the test set is insufficient to produce a statistically significant drop in accuracy that would indicate the mutant being killed.

\textbf{Experimental Design.} 
\begin{figure*}[t]
    \centering
    \includegraphics[width=\textwidth]{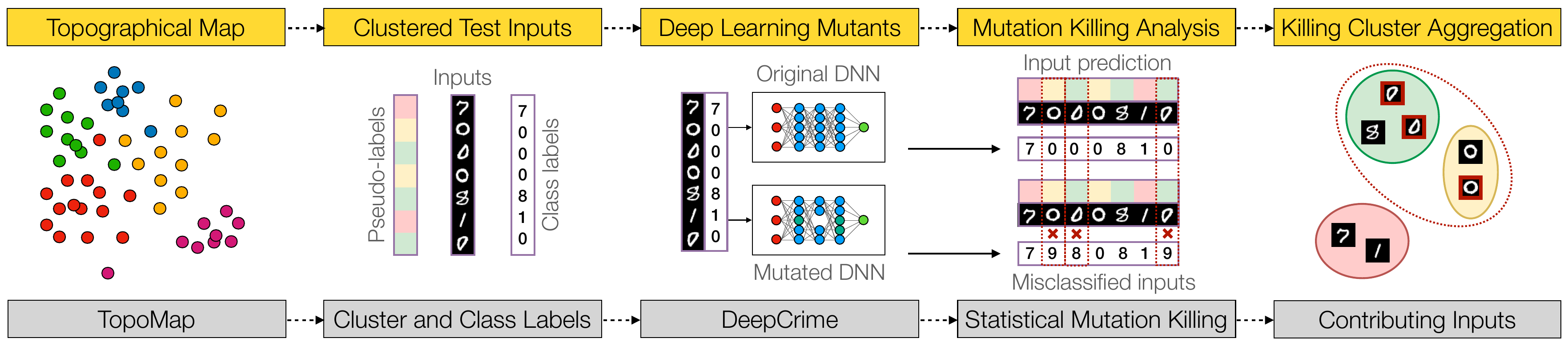}
    \caption{Structure of the mutation-based evaluation pipeline}
    \label{fig:pipeline2}
\end{figure*}
The experimental procedure is depicted in Fig.~\ref{fig:pipeline2}.
After computing a topographical map of the inputs using \topomap (Step 1), we assign each available test input to a \topomap cluster (Step 2). The inputs used for evaluation are drawn from the test dataset available in each application domain (e.g., digit recognition). These inputs are considered in their raw, pre-embedding form (e.g., image pixels), together with their original ground-truth labels (e.g., the digit to be recognised).

In Step 3 (deep learning mutants), the inputs are fed to a set of pre-trained original DL models, as well as to their corresponding mutants obtained by applying different mutation operators and configurations. Multiple instances of the pre-trained original DL models are required to support the statistical computation of mutation killing used in the subsequent step.
%To obtain mutants, we use the DeepCrime~\cite{10.1145/3460319.3464825} pre-training DL mutation tool, which implements 24 MOs, each configurable by setting the values of its parameters (e.g., the MO ``change learning rate'' is configurable by setting a specific mutated learning rate). 

In Step 4 (mutation killing analysis), to account for the non-determinism affecting the training process of both the original and mutant models, we compare the predictions produced by the mutants with those of the original DL model across 
$N$ experimental runs ($N=20$ in our study), each corresponding to a separate retraining of both the original model and its mutants.
Specifically, we adopt the statistical notion of mutation killing proposed by Jahangirova and Tonella~\cite{9159089,10.1145/3460319.3464825}. Under this definition, a mutant is considered \textit{killed} if there is a statistically significant difference between the $N$ performance metric values of the original model and those of the mutant (accuracy for classification systems and mean squared error for regression problems), accompanied by a non-negligible and non-small Cohen’s $d$ effect size.

In the last step,
if a mutant is killed, we determine the \emph{contributors}, i.e., the input vectors that potentially contribute to killing the mutant by triggering a misprediction of the mutant. 
Specifically, a contributor is any input test vector for which the probability of being mispredicted ($P^K$) by the mutant instances is higher than that of the original baseline DL model instances. The probability of being mispredicted is calculated across $N$ retrainings (instances) of both the original and the mutated model:
    \begin{eqnarray}
    \textit{Contrib}(M) = \{ x \mid x \in X_\textit{test}\, \land \,P^K_{\textit{DL}}(x) < P^K_{\textit{M}}(x) \} \label{eq:mu_kill1} %\\
    % \textit{Contrib}(M) = \{ x \mid x \in X_\textit{test}\, \land \, \lVert f_{\textit{DL}}(x) - y \rVert < \delta \, \land \lVert f_{\textit{M}}(x) - y \rVert \geq \delta\} \label{eq:mu_kill2} 
    \end{eqnarray}
%\noindent

For regression problems, we introduce an application domain specific threshold $\tau$ on the prediction error to decide if an input is mispredicted, as done in previous work~\cite{10.1145/3460319.3464825}: $\lVert f(x) - y \rVert \geq \tau$, where $f(x)$ is the model prediction and $y$ is the ground-truth value for $x$. In Fig.~\ref{fig:pipeline2} (Step 4), the three misclassified inputs indicated with a red cross (2nd, 3rd, and 7th entries) are contributors.

We sort the clusters produced by our approach in descending order of contributor density and incrementally aggregate them into a candidate killing set, until the obtained aggregation of the clusters results in the mutant being statistically significantly killed~\cite{9159089}. We consider our approach to be effective in killing a mutant if a random selection of inputs with the same size as our killing aggregation has a substantially lower probability of killing the same mutant. In fact, this would indicate that the features characterising each of our clusters are useful to identify a high concentration of killing inputs, which in turn would allow developers to kill a mutant with a small number of instances. We repeated all our experiments 10 times to account for the randomness associated with clustering and random selection of inputs.
%We also compare our approach with DeepHyperion, comparing the density of contributor inputs in the mutation killing map regions, considering the two respective maps of the test inputs.

\textbf{Metrics.} To answer this question, we introduce the following three metrics computed for each mutant $M$ and then aggregated across MOs.
The proportion of input instances gathered within the clusters $K$ in the killing aggregation $\mathcal{A}$ over the size of the entire input test set.
        \begin{equation}
            \rho_k = \dfrac{1}{|X_\textit{test}|} \sum_{K \in \mathcal{A}} |K|
        \end{equation}
The proportion of contributors in the aggregation, computed as the ratio between the total number of contributors $c(K)$ in the aggregated clusters and the size of the aggregation $\mathcal{A}$.
        \begin{equation}
            \rho_c = \dfrac{\sum_{K \in \mathcal{A}} c(K)}{|\mathcal{A}|} 
        \end{equation}
\begin{comment}
The Gini Impurity (GI)~\cite{9420091} with respect to the number $c(\mathcal{A})$ of contributors in the killing aggregation. A low value of GI\changed{, together with a high value of $\rho_c$,} indicates that the cluster aggregation $\mathcal{A}$ contains mostly killing contributors. \nargiz{it can also be that we have a small number of contributors}\gianmarco{That's why I added the sentence in blue, if both conditions are true it means that the cluster aggregation contains mostly killing contributors}%\paolo{It's a bit strange that we do not use a simple density metric, like the proportion of contributors in the aggregation. If feasible, I'd include both density and GI.}\gianmarco{We have that information on the raw results, but I did not put it into the paper to avoid making the tables too heavy. But I can add it.}

    \begin{equation}
        \textit{GI} = 1 - \left( \dfrac{c(\mathcal{A})}{|\mathcal{A}|} \right)^2 - \left( 1 - \dfrac{c(\mathcal{A})}{|\mathcal{A}|} \right)^2\\
    \end{equation}
\end{comment}

We check whether a random selection of a number of inputs equal to the size of the killing aggregation $\mathcal{A}$ is capable of killing the mutant. For that purpose, we compute the probability $\rho_d$ that a mutant $M$ is killed across $R$ random input selections of size $|\mathcal{A}|$.
    \begin{equation}
    \rho_d = \dfrac{1}{R} \displaystyle \sum_{i=1}^R d_i,\quad \text{with } 
    d_i =
        \begin{cases}
            1, & M\, \text{is killed by}\, |\mathcal{A}|\, \text{random inputs}\\
            0, & \text{otherwise}
        \end{cases}
    \end{equation}\\

\begin{comment}
    We want to assess whether there exists an aggregation of a strict subset of all the clusters able to kill the mutant. For that purpose, we compute the probability $\rho_d$ that a mutant is killed through our experiments.
    \begin{equation}
    \rho_d(M) = \dfrac{1}{\textit{tot}_\textit{exp}} \displaystyle \sum_{\textit{exp}} d,\quad \text{with } 
    d =
        \begin{cases}
            1, & M\text{ is killed} \land 0 < |\mathcal{A}| < \sum_K |K|\\
            0, & \text{otherwise}
        \end{cases}
    \end{equation}
\end{comment}

\textbf{RQ4 (Comparison with DeepHyperion):} \emph{How does the map produced by \topomap differ from that produced by the state-of-the-art technique DeepHyperion?}

Among the subjects considered in this study, the current implementation of DeepHyperion is applicable only to the MNIST dataset, forcing us to constrain our comparison to this subject. The other subject, BeamNG, used in the original DeepHyperion study, is aimed at system-level testing. The test units for this subject are driving trajectories used by the BeamNG simulator. These trajectories are not directly fed to a neural model operating the autonomous vehicle, but are rather influencing the simulation scenario that produces track images that are consequently passed to the model. Such a setup is not compatible with our approach that operates directly on a model's training and test data.

To compare our approach to DeepHyperion on the MNIST subject, we first assess how the maps produced by the two tools relate to each other, specifically whether they exhibit complementarity or overlap. One way to quantify this relationship is to compute the similarity between the map generated by \topomap and that produced by DeepHyperion.

DeepHyperion is designed to operate at the label level, partitioning the input space for each class in a dataset separately. DeepHyperion operates based on a set of manually defined features intended to capture input characteristics that are relevant to deep learning model predictions and that correlate with an input’s fault-revealing capability. Initial output of DeepHyperion represents a set of feature values assigned to each input. The inputs can be then arranged into a grid-like structure, called a feature map, that is constructed by binning the range of values of each feature into $b$ bins. Each cell of such a map corresponds to a specific combination of features and is populated by inputs with features values falling into corresponding bins. 

\textbf{Experimental Design.} To ensure a fair comparison, we carry out the similarity analysis on a per-label basis, treating the map generated by our method accordingly. Specifically, for each label, we construct a label-specific map by isolating the clusters populated by inputs belonging to that label, while ignoring all others. We then count the number of clusters in the resulting map and configure the DeepHyperion cells (i.e., select the number of bins $b$) such that the number of populated cells is as close as possible to the number of clusters in \textsc{TopoMap}’s map. Finally, we compute similarity scores for each label and aggregate them across all labels for a final similarity measure. 
%We then aggregate all resulting clusters or cells into unified maps, perform the similarity comparison at the level of individual clusters or cells, and finally aggregate the similarity scores across all clusters or cells.

Furthermore, to compare the tools' ability to discover fault-exposing inputs, we carry out the mutation analysis described in RQ3, treating DeepHyperion cells as clusters.
%the procedure illustrated in Fig.~\ref{fig:pipeline2} to DeepHyperion cells \nargiz{if this figure and procedure were explained before, I would refer to them here}\gianmarco{we explained them before in the approach section as for the evaluation of TopoMap using mutation analysis, but we also refer to them in RQ3, in which TopoMap is compared against the random selection} \nargiz{yeah, what is mean is, can you please put smth like (see Section \ref{sec:approach})}\gianmarco{Ohh, I see, thanks! At this point, I would rephrase the sentence}. 
Following the approach adopted for RQ3, we then identify aggregations of cells and assess the number of killing inputs contained within them.

Finally, we also evaluate the ability of each technique to group together critical inputs and separate them from the other ones. We deem a test input to be critical either if its classification is subject to high uncertainty by the model under test or if it is misclassified. Hence, for this analysis we collect different architectures aimed at solving the MNIST digit recognition task from the relevant literature and online repositories. For each model under test, we assess whether an input is misclassified by the model, qualifying as a \emph{failing input} %faulty input
or if its prediction shows high-uncertainty. To quantify the uncertainty w.r.t. a test input, we rely on two established metrics~\cite{MPP}, namely DeepGini and Vanilla Softmax and we set a threshold deeming uncertainties above the $95^\textit{th}$ percentile to be high. After getting this categorisation of the inputs into high-uncertainty/misclassified, we evaluate the impurity of their spread within the maps produced by \topomap and DeepHyperion.

\textbf{Metrics}. To answer this question, we first compare the maps generated by the two approaches by measuring the similarity score~\cite{10304866}, a metric based on the Gini impurity~\cite{9420091}.

Assuming a classification problem involving $\mathcal{L} $ classes, let $A$ and $B$ be two maps of the input space consisting of $K_A$ and $K_B$ clusters, respectively. For a cluster $\mathcal{C}_A^i$ from $A$, %where $i$ is the cluster identifier,
the Gini impurity of $\mathcal{C}_A^i$ with respect to the map $B$ is defined as:
\begin{equation}
    \textit{GI}(\mathcal{C}_A^{i},\, B) = 1 - \displaystyle \sum_{j=1}^{K_B} (p_{Bj})^2 
\label{eq:giniimpurity}
\end{equation}
where $p_{Bj}$ denotes the probability that inputs of cluster $j$ of map $B$ also occur in $\mathcal{C}_A^i$, indicating an overlap between the clusters $i$ and $j$ in the two maps.

By averaging the $\textit{GI}$~\eqref{eq:giniimpurity} values across all clusters in map $A$, using the cluster ids of map $B$ as labels, we obtain the Gini Similarity for a class $\ell \in \mathcal{L}$:
\begin{equation}
    {\textit{GS}(A,\,B)}_{\mathcal{\ell\, \in\, L}} = 1 - \dfrac{1}{K_A} \displaystyle \sum_{i=1}^{K_A} GI(\mathcal{C}_A^i,\, B)
\end{equation}

To obtain the Gini Similarity for the entire map, we then average across class labels in $\mathcal{L}$:
\begin{equation}
    \textit{GS}(A,\,B) = \dfrac{1}{|\mathcal{L}|} \displaystyle \sum_{\mathcal{\ell}=1}^{|\mathcal{L}|} {\textit{GS}(A,\,B)}_{\mathcal{\ell\, \in\, L}}
\end{equation}

Since this similarity measure is inherently asymmetric, as it depends on how clusters from one map are distributed over those of the other, the comparison must account for both directions. To mitigate this asymmetry and avoid bias toward either map, we define the similarity between the two maps as the maximum $\textit{GS}$ obtained in either direction:
\begin{equation}
    \textit{Sim}(A,\,B) = \max \left( \textit{GS}(A,B),\, \textit{GS}(B,A) \right)
\end{equation}

Furthermore, to enable a comparison of the ability of the two approaches to aggregate fault-revealing inputs into specific regions, we use the three metrics introduced in RQ3, namely the proportion $\rho_k$ of test inputs in the aggregation, the density $\rho_c$ of contributing inputs in the cluster aggregation, and the probability $\rho_d$ of the mutant being killed by the aggregation.

For evaluating the impurity of a $K$-cluster/cell map w.r.t. the critical inputs, we employ the standard definition of Gini impurity:
\begin{equation}
    \textit{GI} = \dfrac{1}{K} \displaystyle \sum_{i=1}^{K} \left( 1 - \left( {m_i}^2 + {(1-m_i)}^2 \right)\right)
\end{equation}
with $m_i$ being the probability for an input in a cluster/cell $i$ of the map of being either high-uncertainty input or misclassified.\\

\textbf{RQ5 (Killing Propagation):} \emph{Do different aggregations, which kill different MOs and mutation configurations, repeatedly include the same clusters?}

The goal of this question is to assess whether the obtained killing clusters persist in triggering misbehaviours across different MOs and MO configurations. This would indicate that the failure regions of our topographical map are useful across different types of faults possibly affecting the DL system under test.

\textbf{Experimental Design and Metrics.} To assess this persistence, we measure the \textit{cluster killing strength} as the percentage of aggregations ($\rho_a$) that contain a given cluster when considering different mutants. %The existence of clusters with high killing strength would indicate that their failure exposure capability is not confined to a single, specific type of fault. 
We also measure whether the same aggregations consistently kill at least half or all the configurations of analysed MOs (fault types). The existence of clusters with high killing strength would indicate that their fault-exposure %failure-exposure
capability is not confined to a single, specific type of fault or its configuration.

\subsection{Embedding Technique Selection}
Given the wide range of available embedding techniques, we select a subset of them that covers different assumptions about the structure of the data and provides diverse representation. Rather than focussing the analysis on a specific family of embedding techniques, our selection represents linear and non-linear approaches as well as supervised and unsupervised methods.

In particular, we consider: (i) Principal Component Analysis (PCA)~\cite{da6385d2-9c65-3860-bbcd-b821fdff69ff} that returns a set of components obtained as linear combinations of the original dataset features; (ii) Uniform Manifold Approximation and Projection (UMAP)~\cite{10.1162/neco-a-01434}, which performs a non-linear dimension reduction on the dataset; (iii) t-distributed Stochastic Neighbour Embedding (t-SNE)~\cite{JMLR:v9:vandermaaten08a}, a non-linear dimension reduction method optimised for two or three dimensions; (iv) Singular Value Decomposition (SVD)~\cite{Wall2003}, which projects data onto a lower-dimensional space while retaining the most significant singular values; (v) Linear Discriminant Analysis (LDA)~\cite{Zhao2024}, which identifies the feature subspace by maximising the separation between classes.

Additional embedding techniques are used for selected data scenarios due to input-specific characteristics that limit their applicability across all scenarios: (vi) Isomap~\cite{doi:10.1126/science.290.5500.2319}, a non-linear dimensionality reduction technique aiming at preserving the global geometry of the data; (vii) Autoencoder (AE), a type of a DNN that learns a dimensionally efficient (encoded) representation of the data, which can  be subsequently used by other methods or eventually decoded back into a form that is roughly similar to the original; (viii) Word2Vec~\cite{mikolov2013efficientestimationwordrepresentations}, a model that produces vector representations of words that preserve their proximity to neighbours within a corpus of text.

\subsection{Clustering Algorithm Selection}

In our study, we analyse the structure of the obtained embeddings using several clustering algorithms to identify groups of semantically related inputs. Since different clustering algorithms make different assumptions about the structure and distribution of the data, relying on a single approach may favour a particular type of cluster structure. We therefore consider representative algorithms from different clustering families, namely K-means~\cite{https://doi.org/10.1348/000711005X48266}, BIRCH~\cite{10.1145/233269.233324}, HDBSCAN~\cite{10.1007/978-3-642-37456-2_14}, and affinity propagation~\cite{doi:10.1126/science.1136800}. These algorithms provide complementary clustering strategies: K-means represents flat clustering, BIRCH provides a hierarchical approach designed for efficient processing of larger datasets, HDBSCAN identifies density-based structures and can distinguish noise from clustered samples, while affinity propagation uses an exemplar-based approach. 

\section{Results}\label{sec:results}
    
\textbf{RQ1. Discrimination:} \emph{How useful is the proposed approach to select the optimal clustering configuration, i.e., the clustering configuration that better discriminates the test input instances?}

Table~\ref{raw_emb_components} shows the size of the embedded feature space for each considered data scenario. It can be seen that the data dimensionality is significantly reduced: more than 80\% of the original size in all cases. Besides the expected outcome of enhancing significant, discriminative features, the reduced size of the feature space has a positive impact on the computational workload of the pipeline generating the topographical map. 

    \begin{table}[t]
    \caption{Size of the test set and number of components in the feature space in their original raw form compared to the reduced dimensionality in the embedded form for each data scenario. Although both UD and UE datasets concern regression problems, we can group instances into \emph{buckets} for the sake of having a starting point with respect to the number of clusters for the generation of the topographical map.}\label{raw_emb_components}%
    \small
    \centering
    \begin{tabular}{@{}lrrrrr@{}}
    \toprule
    \textbf{Data} 
    & \textbf{\#Inputs} 
    & \textbf{\#Classes} 
    & \textbf{Data dim.} 
    & \textbf{Embedding dim.}
    & \textbf{Retained dim.} \\
    \midrule
    MN & 10,000 & 10 & 784   & 86 & 10.97\% \\
    CF & 10,000 & 10 & 3,072  & 98 & 3.19\% \\
    SR & 1,350  & 5  & 8,000  & 565 & 7.06\% \\
    RE & 2,246  & 46 & 10,000 & 1,910 & 19.10\% \\
    UD & 2,432  & (\textit{3}) & 51,200 & 274 & 0.53\% \\
    UE & 25,857 & (\textit{9}) & 2,162  & 35 & 1.62\% \\
    \bottomrule
    \end{tabular}
    \end{table}

Table~\ref{Pipeline1Results} shows the results of the application of different combinations of embeddings and clustering algorithms. The row highlighted in grey reports the minimum weighted accuracy of the best-performing topographical map, chosen among all considered configurations of embeddings and clustering algorithms (the row `\emph{$\min w_{acc}^{A,B}$}'). For each subject system, the best configuration is reported in the rows `\emph{Embedding}' and `\emph{Clustering}', while the row `\emph{$k^*$}' shows the optimal number of clusters identified for each corresponding configuration. 

Results indicate that none of the commonly employed metrics like Silhouette score (the row `\emph{Sil. score}'), the Davies-Bouldin (the row `\emph{DB index}') or the Calinski–Harabasz (the row `\emph{CH index}') indices for cluster evaluation is consistently correlated with the selection of the best clustering configuration performed by the proposed approach (the row `\emph{$\min w_{acc}^{A,B}$}'), based on a DNN classifier. 

%The first thing to be noticed is the fact that none of the commonly employed metrics for cluster evaluation is consistently correlated in any way with the selected topographical map. 
    \begin{table}[t]
    \caption{Selected topographical map for each data scenario by embedding/clustering algorithm configuration with respect to the accuracy of the cluster pair most difficult to discriminate by a DNN. Widely used metrics for cluster quality assessment (Silhouette score, DB/CH index) are  included for comparison.}\label{Pipeline1Results}%
    \small
    \centering
    \begin{tabular}{lrrrrrrc@{}}
    \toprule
    & \textbf{MN} & \textbf{CF} & \textbf{SR} & \textbf{RE} & \textbf{UD} & \textbf{UE} \\
    \midrule
    \textbf{Embedding}   & SVD   & PCA   & t-SNE & PCA   & t-SNE & LDA \\
    \textbf{Clustering} & K-means & K-means & K-means & BIRCH & BIRCH & K-means \\
    $\bm{k^*}$           & 90    & 40    & 40    & 138   & 9     & 27 \\
    \midrule
    \textbf{Sil. score$^{(\uparrow)}$}
                        & 0.065 & 0.023 & 0.376 & -0.088 & 0.380 & 0.120 \\
    \textbf{DB index}$^{(\downarrow)}$
                        & 2.453 & 2.806 & 0.806 & 3.607 & 0.820 & 1.677 \\
    \textbf{CH index}$^{(\uparrow)}$
                        & 120.077 & 246.947 & 2,555.596 & 6.164 & 2,422.404 & 2,057.630 \\
   \cellcolor[gray]{0.85}{$\bm{\min w_{acc}^{A,B}}{}^{(\uparrow)}$}
                        & \cellcolor[gray]{0.85}{0.941}
                        & \cellcolor[gray]{0.85}{0.919}
                        & \cellcolor[gray]{0.85}{0.439}
                        & \cellcolor[gray]{0.85}{0.679}
                        & \cellcolor[gray]{0.85}{0.994}
                        & \cellcolor[gray]{0.85}{0.920} \\
    $\bm{w_{acc}^{A,B}}$
                        & 0.999 & 0.992 & 0.982 & 0.996 & 0.998 & 0.986 \\
    \bottomrule
    \end{tabular}
    \end{table}

The high values (higher than $90\%$) of the cumulative average weighted pairwise accuracy (the row `\emph{$w_{acc}^{A,B}$}')  suggest that the trained DNN model is always able to successfully recognise the generated clusters. However, we can also see that the lowest weighted pairwise accuracy (the row `\emph{${\min w_{acc}^{A,B}}$}') changes significantly across the subjects and configurations (see the replication package~\cite{github:repo} for full results). In fact, it can be observed that the choice of the best-performing configuration is dependent on the application scenario, with no specific configuration clearly dominating the others across all considered subjects. Moreover, the choice of a specific embedding (e.g., for UD) or clustering algorithm (e.g., for MN) can have a significant impact on the accuracy (see replication package~\cite{github:repo} for full results) and, as a result, produce very different topographical maps. In particular, we can see that the combination of a linear embedding with K-means is the %best
choice achieving best cluster separation in 3 cases out of 6, while for the remaining 3 subjects, the selected topographical map is produced either by BIRCH or by a non-linear embedding. The optimal number of clusters also varies from dataset to dataset, ranging from 9 for UD to 138 for the RE dataset.

    \begin{custombox}[Answer to RQ1] 
    Our results show that, for each subject, \topomap can generate a map with distinguishable clusters, and that the adopted DNN-based evaluation pipeline is important to discriminate maps of different cohesiveness quality, because different subjects require different clustering configurations.
    %For each data scenario, we obtain at least one topographical map where clusters are recognisable and distinguishable, thus we can conclude that those clusters successfully group together and separate different instances by some identifiable discriminative features.
    \end{custombox}
\textbf{RQ2. Human Alignment}: \emph{How does the cohesion and separability evaluation of the DNN assessor align with that of humans?}

% For each of the four  questionnaires, we recruited 20 participants. For each questionnaire, we considered only the responses that passed the attention check, republishing the questionnaire until the expected number of responses was obtained. Hence, we have 20 respondents for each of the two questionnaires on the MN subject and 20 respondents for each of the two questionnaires on the CF subject. The results, aggregated by subject, are reported in Table~\ref{table:results_human_study}.

% We constructed four questionnaires, two for each of the selected subjects. In total, we collected 20 responses for each questionnaire. 
%We based the evaluation on two image-based classification subjects: MN and CF. For each subject, we created two questionnaires and recruited 20 respondents for each one.

For each selected image classification subject (MN and CF), we constructed two questionnaires and collected 20 responses per questionnaire. The results, aggregated by subject, are reported in Table~\ref{table:results_human_study}. For the statistics, we report 3 times the standard error (3SE) corresponding to a $99.73\%$ confidence level.

\begin{table}[b]
\caption{PCC, mean and 3 times the standard error (3SE) computed between human-based metrics and DNN-based metrics with respect to classification according to the cluster labels. Highlighted correlation coefficient indicates statistical significance ($p <$ 0.05).}\label{table:results_human_study}%
\small
\centering
\setlength{\tabcolsep}{4pt}
\renewcommand{\arraystretch}{1.3}
\begin{tabular}{@{}ccrccrccrc@{}}
\toprule
%    & \multicolumn{6}{c}{\textbf{Data}}\\
%    \cmidrule{2-7}
& & \multicolumn{2}{c}{$\bm{w_{acc}^{A,B}}$} & & \multicolumn{2}{c}{$\bm{t_{acc}^{A,B}}$} & & \multicolumn{2}{c}{$\bm{{t'}_{acc}^{A,B}}$}\\ 
\cmidrule{3-4}\cmidrule{6-7}\cmidrule{9-10}
& \textbf{Metric} & \multicolumn{1}{c}{\textbf{PCC}} & \multicolumn{1}{c}{\textbf{Mean $\bm{\pm}$ 3SE}} & & \multicolumn{1}{c}{\textbf{PCC}} & \multicolumn{1}{c}{\textbf{Mean $\bm{\pm}$ 3SE}} & & \multicolumn{1}{c}{\textbf{PCC}} & \multicolumn{1}{c}{\textbf{Mean $\bm{\pm}$ 3SE}}\\
\midrule
\multirow{2}{*}{\textbf{MN}} & $h_{\textit{acc}}^{A,B}$  &  \cellcolor[gray]{0.85}{0.735} & \multirow{2}{*}{0.915 $\pm$ 0.043} & &  \cellcolor[gray]{0.85}{0.736} & \multirow{2}{*}{0.919 $\pm$ 0.041} & &  \cellcolor[gray]{0.85}{0.622}  & \multirow{2}{*}{0.913 $\pm$ 0.053}\\
                    & $h_{\textit{diff}}^{A,B}$ &  \cellcolor[gray]{0.85}{-0.691} &   & & \cellcolor[gray]{0.85}{-0.631} &  & & \cellcolor[gray]{0.85}{-0.585} & \\
\midrule
\multirow{2}{*}{\textbf{CF}} & $h_{\textit{acc}}^{A,B}$  &  0.351 &  \multirow{2}{*}{0.856 $\pm$ 0.079}   & &  \cellcolor[gray]{0.85}{0.401} & \multirow{2}{*}{0.882 $\pm$ 0.058} & &  0.347  & \multirow{2}{*}{0.866 $\pm$ 0.080}\\
                    & $h_{\textit{diff}}^{A,B}$ &  0.086 &  & &  0.050 &  & & -0.050  & \\
%\midrule
%\multirow{2}{*}{\textbf{UE}} & $h_{\textit{acc}}^{A,B}$  &\\
%                    & $h_{\textit{diff}}^{A,B}$ &\\
\bottomrule
\end{tabular}
\end{table}

Within the MN data scenario, empirical evidence suggests a strong correlation for both $h_{\textit{acc}}^{A,B}$ and $h_{\textit{diff}}^{A,B}$ with all the DNN-based metrics. In particular, the negative correlation between $h_{\textit{diff}}^{A,B}$ and the other metrics suggests that the difficulty for a human assessor to distinguish a pair of clusters is aligned with the inaccuracies of the DNN model. However, on the CF dataset, results show a less marked relationship between the metrics, with a moderate correlation between $h_{\textit{acc}}^{A,B}$ and the DNN-based metrics. In the MN scenario, the measured correlation between $w_{acc}^{A,B}$ and $h_{\textit{acc}}^{A,B}$ achieves a low $p$-value ($p < 0.05$) thus suggesting statistical significance. The same also occurs with the $h_{\textit{diff}}^{A,B}$ metric. On the other hand, in the CF scenario, we observe a moderate correlation between the DNN-based metrics and  $h_{\textit{acc}}^{A,B}$. However, only the correlation with $t_{acc}^{A,B}$ achieves statistical significance, while the $p$-value of  PCC between $h_{acc}^{A,B}$ and $w_{acc}^{A,B}$ is 0.057, and 0.061 with respect to ${t'}_{acc}^{A,B}$. Conversely, none of the correlations computed with respect to $h_{\textit{diff}}^{A,B}$ achieves statistical significance. %\changed{Finally, for the UE scenario...}

When performing the classification tasks, assessors had also the possibility to indicate which strategy they followed for assigning images to a cluster. In the case of the MN questionnaires, the \emph{digit value} was taken into account by 73\% of the participants, \emph{digit orientation} by 58\%, while \emph{digit boldness} by 65\%. On the other hand, for the CF questionnaires, \emph{image subject} was considered by 83\% of the participants, while \emph{image colour} was deemed relevant by 60\%. %\changed{Instead, in the UE questionnaires, XX\% of the participants declared to have considered the \emph{eye shape} in their decision process, XX\% the \emph{gaze direction}, while XX\% considered the \emph{brightness} of the image.}

%In order to uncover a potential, implicit, bias, we asked the assessors to declare whether they assumed that for each task, the split across clusters was exactly of 5 elements per clusters. In the MN data scenario, 82\% of the participants answered \emph{yes}, while in the CF scenario, 80\% made this assumption.

Despite meeting our expectations, the human study comes with some limitations that should be taken into account. First of all, as for the CF dataset,  images depict real world subjects with a very low resolution, making it hard even for a human  to understand what is represented (this observation came out during a pilot of our study). This is backed by the fact that, by considering solely the human-based metrics, we get that $h_{acc}^{A,B}$ and $h_{\textit{diff}}^{A,B}$ are moderately negatively correlated (-0.634) in a statistically significant way ($p$-value: \num{1.7e-4}) in the MN scenario, while for the questionnaires involving CF, the assessment performed by the participants results in a non-discernible negative correlation (-0.288) which does not achieve statistical significance ($p$-value: 0.12). Second, human-based metrics are compared with DNN-inferred metrics even though the two evaluators (human assessor and DNN model) have been supplied different data samples for training. Indeed, for practical reasons, the human assessor has been shown only a random selection of inputs of a cluster to give them an idea of the features represented by such cluster. Instead, the DNN model was provided with the entire training set for that cluster. This might have affected to some degree the human assessors' performance in the tasks related to the CF scenario, whose clusters are generally heterogeneous with respect to the original class label, the underlying features being not necessarily easily recognisable by humans.

\begin{custombox}[Answer to RQ2] 
Empirical results show correlation between human-measured accuracy and accuracy measured through DNNs, as well as some degree of negative correlation between the human-perceived task difficulty and the DNN accuracies, particularly on the MN dataset. However, the dataset with more complex features (CF) shows that humans and DNN  might consider different features when discriminating input clusters. 
\end{custombox}
% \textbf{RQ3. Fault Detection:} \emph{Are clusters indicative of failure regions in DL systems?}
\textbf{RQ3. Fault Detection:} \emph{Are clusters indicative of faults in DL systems?}

\begin{table*}[t]
    \caption{Proportion $\rho_k$ of input instances included into the killing cluster aggregation for three classification data scenarios (MN, CF and SR) averaged over ten empirical runs. The density of contributing inputs $\rho_c$ quantifies the concentration of killing inputs in the considered cluster aggregations. Results are compared with the concentration of contributors $\rho_c$ and the probability $\rho_d$ of the mutant being killed by the aggregation obtained from the feature regions identified by a random selection.}\label{Pipeline2ResultsA}%
    \resizebox*{\textwidth}{!}{
    \begin{tabular}{@{}ccccccccccccccccccccccccccccr@{}}
    \toprule
    % Table column names
    & & \multicolumn{8}{c}{\textbf{MN}} & & \multicolumn{8}{c}{\textbf{CF}} & & \multicolumn{8}{c}{\textbf{SR}}\\
    \cmidrule{3-10}\cmidrule{12-19}\cmidrule{21-28}
    & \textbf{MO} & \textbf{\#} & $\bm{t_c}$ & & $\bm{\rho_\bm{k}}$ & $\bm{\rho_\bm{c}}$ & $\bm{\rho^{(R)}_\bm{c}}$ & $\bm{\rho_d}$ & $\bm{\rho^{(R)}_d}$ & & \textbf{\#} & $\bm{t_c}$ & & $\bm{\rho_\bm{k}}$ & $\bm{\rho_\bm{c}}$ & $\bm{\rho^{(R)}_\bm{c}}$ & $\bm{\rho_d}$ & $\bm{\rho^{(R)}_d}$  & & \textbf{\#} & $\bm{t_c}$ & & $\bm{\rho_\bm{k}}$ & $\bm{\rho_\bm{c}}$ & $\bm{\rho^{(R)}_\bm{c}}$ & $\bm{\rho_d}$ & $\bm{\rho^{(R)}_d}$ \\
    \midrule
    \parbox[t]{2mm}{\multirow{15}{*}{\rotatebox[origin=c]{90}{\textit{Killable}}}}
& TAN & 2 & 0.014 & & 0.025 & \cellcolor[gray]{0.85}{0.070} & 0.016 & \cellcolor[gray]{0.85}{1.000} & 0.300  & &4 & 0.462 & & 0.027 & \cellcolor[gray]{0.85}{0.541} & 0.465 & \cellcolor[gray]{0.85}{1.000} & 0.575  & &--- & --- & & --- & --- & --- & --- & --- \\
& RAW & 3 & 0.013 & & 0.030 & \cellcolor[gray]{0.85}{0.060} & 0.013 & \cellcolor[gray]{0.85}{1.000} & 0.233  & &12 & 0.433 & & 0.053 & \cellcolor[gray]{0.85}{0.511} & 0.427 & \cellcolor[gray]{0.85}{1.000} & 0.750  & &--- & --- & & --- & --- & --- & --- & --- \\
& ACH & 9 & 0.019 & & 0.032 & \cellcolor[gray]{0.85}{0.096} & 0.018 & \cellcolor[gray]{0.85}{1.000} & 0.322  & &5 & 0.573 & & 0.019 & \cellcolor[gray]{0.85}{0.653} & 0.576 & \cellcolor[gray]{0.85}{1.000} & \cellcolor[gray]{0.85}{1.000}  & &1 & 0.035 & & 0.034 & \cellcolor[gray]{0.85}{0.311} & 0.029 & \cellcolor[gray]{0.85}{1.000} & 0.000 \\
& HBS & 3 & 0.015 & & 0.022 & \cellcolor[gray]{0.85}{0.075} & 0.016 & \cellcolor[gray]{0.85}{1.000} & 0.033  & &1 & 0.391 & & 0.033 & \cellcolor[gray]{0.85}{0.466} & 0.375 & \cellcolor[gray]{0.85}{1.000} & 0.100  & &--- & --- & & --- & --- & --- & --- & --- \\
& RCP & --- & --- & & --- & --- & --- & --- & ---  & &--- & --- & & --- & --- & --- & --- & ---  & &4 & 0.109 & & 0.039 & \cellcolor[gray]{0.85}{0.548} & 0.113 & \cellcolor[gray]{0.85}{1.000} & 0.450 \\
& HNE & 3 & 0.023 & & 0.028 & \cellcolor[gray]{0.85}{0.107} & 0.028 & \cellcolor[gray]{0.85}{1.000} & 0.433  & &3 & 0.459 & & 0.208 & \cellcolor[gray]{0.85}{0.522} & 0.453 & \cellcolor[gray]{0.85}{1.000} & 0.867  & &4 & 0.178 & & 0.033 & \cellcolor[gray]{0.85}{0.557} & 0.162 & \cellcolor[gray]{0.85}{1.000} & 0.525 \\
& TCL & 6 & 0.043 & & 0.031 & \cellcolor[gray]{0.85}{0.332} & 0.043 & \cellcolor[gray]{0.85}{1.000} & 0.367  & &4 & 0.416 & & 0.035 & \cellcolor[gray]{0.85}{0.556} & 0.416 & \cellcolor[gray]{0.85}{1.000} & 0.750  & &6 & 0.199 & & 0.033 & \cellcolor[gray]{0.85}{0.637} & 0.190 & \cellcolor[gray]{0.85}{1.000} & 0.667 \\
& HLR & 3 & 0.098 & & 0.021 & \cellcolor[gray]{0.85}{0.308} & 0.089 & \cellcolor[gray]{0.85}{1.000} & 0.367  & &--- & --- & & --- & --- & --- & --- & ---  & &--- & --- & & --- & --- & --- & --- & --- \\
& LCH & 10 & 0.028 & & 0.017 & \cellcolor[gray]{0.85}{0.153} & 0.027 & \cellcolor[gray]{0.85}{1.000} & 0.640  & &--- & --- & & --- & --- & --- & --- & ---  & &12 & 0.999 & & 0.014 & \cellcolor[gray]{0.85}{1.000} & \cellcolor[gray]{0.85}{1.000} & \cellcolor[gray]{0.85}{1.000} & \cellcolor[gray]{0.85}{1.000} \\
& OCH & 3 & 0.028 & & 0.019 & \cellcolor[gray]{0.85}{0.132} & 0.026 & \cellcolor[gray]{0.85}{1.000} & 0.333  & &4 & 0.347 & & 0.095 & \cellcolor[gray]{0.85}{0.415} & 0.346 & \cellcolor[gray]{0.85}{1.000} & 0.625  & &4 & 0.079 & & 0.029 & \cellcolor[gray]{0.85}{0.439} & 0.069 & \cellcolor[gray]{0.85}{1.000} & 0.425 \\
& CWI & 4 & 0.453 & & 0.024 & \cellcolor[gray]{0.85}{0.534} & 0.444 & \cellcolor[gray]{0.85}{1.000} & 0.625  & &3 & 0.882 & & 0.021 & \cellcolor[gray]{0.85}{0.913} & 0.880 & \cellcolor[gray]{0.85}{1.000} & \cellcolor[gray]{0.85}{1.000}  & &1 & 0.044 & & 0.037 & \cellcolor[gray]{0.85}{0.338} & 0.042 & \cellcolor[gray]{0.85}{1.000} & 0.100 \\
& TRD & 4 & 0.114 & & 0.022 & \cellcolor[gray]{0.85}{0.318} & 0.106 & \cellcolor[gray]{0.85}{1.000} & 0.575  & &5 & 0.512 & & 0.025 & \cellcolor[gray]{0.85}{0.579} & 0.507 & \cellcolor[gray]{0.85}{1.000} & 0.560  & &6 & 0.187 & & 0.075 & \cellcolor[gray]{0.85}{0.566} & 0.202 & \cellcolor[gray]{0.85}{1.000} & 0.517 \\
& HDB & 1 & 0.015 & & 0.020 & \cellcolor[gray]{0.85}{0.094} & 0.014 & \cellcolor[gray]{0.85}{1.000} & 0.200  & &1 & 0.428 & & 0.022 & \cellcolor[gray]{0.85}{0.513} & 0.432 & \cellcolor[gray]{0.85}{1.000} & \cellcolor[gray]{0.85}{1.000}  & &--- & --- & & --- & --- & --- & --- & --- \\
& TCO & 1 & 0.122 & & 0.023 & \cellcolor[gray]{0.85}{0.972} & 0.130 & \cellcolor[gray]{0.85}{1.000} & \cellcolor[gray]{0.85}{1.000}  & &1 & 0.419 & & 0.037 & \cellcolor[gray]{0.85}{0.559} & 0.414 & \cellcolor[gray]{0.85}{1.000} & \cellcolor[gray]{0.85}{1.000}  & &6 & 0.213 & & 0.030 & \cellcolor[gray]{0.85}{0.703} & 0.217 & \cellcolor[gray]{0.85}{1.000} & 0.733 \\
& ARM & 4 & 0.260 & & 0.024 & \cellcolor[gray]{0.85}{0.304} & 0.261 & \cellcolor[gray]{0.85}{1.000} & 0.425  & &--- & --- & & --- & --- & --- & --- & ---  & &--- & --- & & --- & --- & --- & --- & --- \\
& TUD & 6 & 0.037 & & 0.044 & \cellcolor[gray]{0.85}{0.229} & 0.040 & \cellcolor[gray]{0.85}{1.000} & 0.450  & &3 & 0.429 & & 0.021 & \cellcolor[gray]{0.85}{0.612} & 0.431 & \cellcolor[gray]{0.85}{1.000} & 0.600  & &--- & --- & & --- & --- & --- & --- & --- \\
    \midrule
    \multicolumn{2}{l}{\textbf{Mean}} & & & & 0.025   & \cellcolor[gray]{0.85}{\underline{0.252}}   &  0.085 & \cellcolor[gray]{0.85}{\underline{1.000}} & 0.420 & & & & &  0.050   & \cellcolor[gray]{0.85}{\underline{0.570}}   & 0.477 & \cellcolor[gray]{0.85}{\underline{1.000}}  & 0.736  & & & & & 0.036   & \cellcolor[gray]{0.85}{\underline{0.567}}  & 0.225 & \cellcolor[gray]{0.85}{\underline{1.000}} & 0.491 \\
    \midrule
    \parbox[t]{2mm}{\multirow{12}{*}{\rotatebox[origin=c]{90}{\textit{Initially Non-Killable}}}}
& AAL & --- & --- & & --- & --- & --- & --- & ---  & &--- & --- & & --- & --- & --- & --- & ---  & &3 & 0.032 & & 0.092 & \cellcolor[gray]{0.85}{0.216} & 0.035 & \cellcolor[gray]{0.85}{0.967} & 0.067 \\
& TAN & 2 & 0.011 & & 0.081 & \cellcolor[gray]{0.85}{0.051} & 0.011 & \cellcolor[gray]{0.85}{1.000} & 0.050  & &2 & 0.361 & & 0.045 & \cellcolor[gray]{0.85}{0.435} & 0.355 & \cellcolor[gray]{0.85}{1.000} & 0.050  & &--- & --- & & --- & --- & --- & --- & --- \\
& RAW & --- & --- & & --- & --- & --- & --- & ---  & &1 & 0.356 & & 0.040 & \cellcolor[gray]{0.85}{0.439} & 0.349 & \cellcolor[gray]{0.85}{1.000} & 0.100  & &--- & --- & & --- & --- & --- & --- & --- \\
& ACH & --- & --- & & --- & --- & --- & --- & ---  & &--- & --- & & --- & --- & --- & --- & ---  & &3 & 0.026 & & 0.058 & \cellcolor[gray]{0.85}{0.175} & 0.025 & \cellcolor[gray]{0.85}{1.000} & 0.067 \\
& RCP & --- & --- & & --- & --- & --- & --- & ---  & &--- & --- & & --- & --- & --- & --- & ---  & &2 & 0.043 & & 0.324 & \cellcolor[gray]{0.85}{0.183} & 0.042 & \cellcolor[gray]{0.85}{0.800} & 0.000 \\
& HNE & 2 & 0.011 & & 0.176 & \cellcolor[gray]{0.85}{0.040} & 0.011 & \cellcolor[gray]{0.85}{0.950} & 0.100  & &1 & 0.336 & & 1.000 & \cellcolor[gray]{0.85}{0.336} & \cellcolor[gray]{0.85}{0.336} & \cellcolor[gray]{0.85}{0.000} & \cellcolor[gray]{0.85}{0.000}  & &--- & --- & & --- & --- & --- & --- & --- \\
& TCL & 1 & 0.010 & & 0.022 & \cellcolor[gray]{0.85}{0.081} & 0.010 & \cellcolor[gray]{0.85}{1.000} & 0.000  & &1 & 0.363 & & 0.141 & \cellcolor[gray]{0.85}{0.415} & 0.358 & \cellcolor[gray]{0.85}{1.000} & 0.200  & &--- & --- & & --- & --- & --- & --- & --- \\
& HLR & 2 & 0.010 & & 0.317 & \cellcolor[gray]{0.85}{0.032} & 0.011 & \cellcolor[gray]{0.85}{0.750} & 0.050  & &2 & 0.337 & & 0.282 & \cellcolor[gray]{0.85}{0.393} & 0.338 & \cellcolor[gray]{0.85}{0.900} & 0.150  & &--- & --- & & --- & --- & --- & --- & --- \\
& CWI & --- & --- & & --- & --- & --- & --- & ---  & &--- & --- & & --- & --- & --- & --- & ---  & &1 & 0.022 & & 1.000 & \cellcolor[gray]{0.85}{0.022} & \cellcolor[gray]{0.85}{0.022} & \cellcolor[gray]{0.85}{0.000} & \cellcolor[gray]{0.85}{0.000} \\
& TRD & 3 & 0.012 & & 0.317 & \cellcolor[gray]{0.85}{0.033} & 0.012 & \cellcolor[gray]{0.85}{0.833} & 0.000  & &3 & 0.360 & & 0.060 & \cellcolor[gray]{0.85}{0.434} & 0.368 & \cellcolor[gray]{0.85}{1.000} & 0.167  & &--- & --- & & --- & --- & --- & --- & --- \\
& TCO & 2 & 0.011 & & 0.093 & \cellcolor[gray]{0.85}{0.047} & 0.012 & \cellcolor[gray]{0.85}{1.000} & 0.100  & &1 & 0.379 & & 0.047 & \cellcolor[gray]{0.85}{0.465} & 0.390 & \cellcolor[gray]{0.85}{1.000} & 0.200  & &--- & --- & & --- & --- & --- & --- & --- \\
& TUD & 1 & 0.013 & & 0.095 & \cellcolor[gray]{0.85}{0.051} & 0.013 & \cellcolor[gray]{0.85}{1.000} & 0.000  & &3 & 0.352 & & 0.236 & \cellcolor[gray]{0.85}{0.419} & 0.342 & \cellcolor[gray]{0.85}{0.800} & 0.033  & &--- & --- & & --- & --- & --- & --- & --- \\
    \midrule
    \multicolumn{2}{l}{\textbf{Mean}} & & & & 0.157  & \cellcolor[gray]{0.85}{\underline{0.048}}   & 0.011 & \cellcolor[gray]{0.85}{\underline{0.933}} & 0.043 & & & & & 0.231   & \cellcolor[gray]{0.85}{\underline{0.418}}    & 0.355 & \cellcolor[gray]{0.85}{\underline{0.837}} & 0.113 & & & & & 0.369   & \cellcolor[gray]{0.85}{\underline{0.150}}  & 0.031 & \cellcolor[gray]{0.85}{\underline{0.692}} & 0.034 \\
    \bottomrule
    \end{tabular}}
    \end{table*}

\begin{table*}[t]
    \caption{Proportion $\rho_k$ of input instances included into the killing cluster aggregation for one classification and two regression data scenarios (RE, UD and UE) averaged over ten empirical runs. The density of contributing inputs $\rho_c$ quantifies the concentration of killing inputs in the considered cluster aggregations. Results are compared with the concentration of contributors $\rho_c$ and the probability $\rho_d$ of the mutant being killed by the aggregation obtained from the feature regions identified by a random selection.}\label{Pipeline2ResultsB}%
    \resizebox*{\textwidth}{!}{
\begin{tabular}{@{}ccccccccccccccccccccccccccccc@{}}
\toprule
& & \multicolumn{8}{c}{\textbf{RE}} & & \multicolumn{8}{c}{\textbf{UD}} & & \multicolumn{8}{c}{\textbf{UE}}\\
\cmidrule{3-10}\cmidrule{12-19}\cmidrule{21-28}
& \textbf{MO} & \textbf{\#} & $\bm{t_c}$ & & $\bm{\rho_\bm{k}}$ & $\bm{\rho_\bm{c}}$ & $\bm{\rho^{(R)}_\bm{c}}$ & $\bm{\rho_d}$ & $\bm{\rho^{(R)}_d}$ & & \textbf{\#} & $\bm{t_c}$ & & $\bm{\rho_\bm{k}}$ & $\bm{\rho_\bm{c}}$ & $\bm{\rho^{(R)}_\bm{c}}$ & $\bm{\rho_d}$ & $\bm{\rho^{(R)}_d}$  & & \textbf{\#} & $\bm{t_c}$ & & $\bm{\rho_\bm{k}}$ & $\bm{\rho_\bm{c}}$ & $\bm{\rho^{(R)}_\bm{c}}$ & $\bm{\rho_d}$ & $\bm{\rho^{(R)}_d}$ \\
\midrule

% ===================== KILLABLE =====================
\parbox[t]{2mm}{\multirow{17}{*}{\rotatebox[origin=c]{90}{\textit{Killable}}}}
& AAL & --- & --- & & --- & --- & --- & --- & ---  & &--- & --- & & --- & --- & --- & --- & ---  & &7 & 0.671 & & 0.035 & \cellcolor[gray]{0.85}{0.911} & 0.674 & \cellcolor[gray]{0.85}{1.000} & 0.943 \\
& TAN & 1 & 0.089 & & 0.995 & \cellcolor[gray]{0.85}{0.089} & 0.089 & \cellcolor[gray]{0.85}{1.000} & \cellcolor[gray]{0.85}{1.000}  & &--- & --- & & --- & --- & --- & --- & ---  & &3 & 0.713 & & 0.035 & \cellcolor[gray]{0.85}{0.801} & 0.716 & \cellcolor[gray]{0.85}{1.000} & \cellcolor[gray]{0.85}{1.000} \\
& RAW & 6 & 0.230 & & 0.001 & \cellcolor[gray]{0.85}{0.742} & 0.204 & \cellcolor[gray]{0.85}{1.000} & 0.317  & &--- & --- & & --- & --- & --- & --- & ---  & &11 & 0.669 & & 0.034 & \cellcolor[gray]{0.85}{0.795} & 0.670 & \cellcolor[gray]{0.85}{1.000} & \cellcolor[gray]{0.85}{1.000} \\
& ACH & 14 & 0.431 & & 0.005 & \cellcolor[gray]{0.85}{0.757} & 0.438 & \cellcolor[gray]{0.85}{1.000} & 0.493  & &--- & --- & & --- & --- & --- & --- & ---  & &5 & 0.793 & & 0.035 & \cellcolor[gray]{0.85}{0.917} & 0.791 & \cellcolor[gray]{0.85}{1.000} & \cellcolor[gray]{0.85}{1.000} \\
& HBS & 3 & 0.098 & & 0.013 & \cellcolor[gray]{0.85}{0.545} & 0.085 & \cellcolor[gray]{0.85}{1.000} & 0.167  & &--- & --- & & --- & --- & --- & --- & ---  & &2 & 0.687 & & 0.036 & \cellcolor[gray]{0.85}{0.879} & 0.684 & \cellcolor[gray]{0.85}{1.000} & \cellcolor[gray]{0.85}{1.000} \\
& HNE & 2 & 0.097 & & 0.125 & \cellcolor[gray]{0.85}{0.472} & 0.081 & \cellcolor[gray]{0.85}{1.000} & 0.250  & &2 & 0.059 & & 0.531 & \cellcolor[gray]{0.85}{0.083} & 0.059 & \cellcolor[gray]{0.85}{1.000} & 0.900  & &4 & 0.571 & & 0.033 & \cellcolor[gray]{0.85}{0.699} & 0.571 & \cellcolor[gray]{0.85}{1.000} & 0.900 \\
& TCL & 3 & 0.217 & & 0.023 & \cellcolor[gray]{0.85}{0.849} & 0.219 & \cellcolor[gray]{0.85}{1.000} & 0.433  & &1 & 0.151 & & 0.234 & \cellcolor[gray]{0.85}{0.258} & 0.154 & \cellcolor[gray]{0.85}{1.000} & \cellcolor[gray]{0.85}{1.000}  & &4 & 0.679 & & 0.035 & \cellcolor[gray]{0.85}{0.785} & 0.679 & \cellcolor[gray]{0.85}{1.000} & \cellcolor[gray]{0.85}{1.000} \\
& HLR & 8 & 0.607 & & 0.002 & \cellcolor[gray]{0.85}{0.944} & 0.609 & \cellcolor[gray]{0.85}{1.000} & 0.588  & &1 & 0.026 & & 0.567 & \cellcolor[gray]{0.85}{0.034} & 0.027 & \cellcolor[gray]{0.85}{1.000} & 0.900  & &3 & 0.505 & & 0.033 & \cellcolor[gray]{0.85}{0.648} & 0.505 & \cellcolor[gray]{0.85}{1.000} & \cellcolor[gray]{0.85}{1.000} \\
& LCH & 9 & 0.146 & & 0.002 & \cellcolor[gray]{0.85}{0.954} & 0.180 & \cellcolor[gray]{0.85}{1.000} & 0.289  & &10 & 0.805 & & 0.121 & \cellcolor[gray]{0.85}{0.814} & 0.805 & \cellcolor[gray]{0.85}{1.000} & 0.990  & &9 & 0.873 & & 0.032 & \cellcolor[gray]{0.85}{0.934} & 0.873 & \cellcolor[gray]{0.85}{1.000} & \cellcolor[gray]{0.85}{1.000} \\
& OCH & 3 & 0.150 & & 0.015 & \cellcolor[gray]{0.85}{0.584} & 0.189 & \cellcolor[gray]{0.85}{1.000} & 0.167  & &3 & 0.064 & & 0.234 & \cellcolor[gray]{0.85}{0.105} & 0.064 & \cellcolor[gray]{0.85}{1.000} & 0.800  & &2 & 0.672 & & 0.034 & \cellcolor[gray]{0.85}{0.838} & 0.672 & \cellcolor[gray]{0.85}{1.000} & \cellcolor[gray]{0.85}{1.000} \\
& CWI & 5 & 0.366 & & 0.188 & \cellcolor[gray]{0.85}{0.734} & 0.319 & \cellcolor[gray]{0.85}{1.000} & 0.580  & &--- & --- & & --- & --- & --- & --- & ---  & &7 & 0.816 & & 0.036 & \cellcolor[gray]{0.85}{0.897} & 0.816 & \cellcolor[gray]{0.85}{1.000} & \cellcolor[gray]{0.85}{1.000} \\
& TRD & 5 & 0.171 & & 0.004 & \cellcolor[gray]{0.85}{0.555} & 0.175 & \cellcolor[gray]{0.85}{1.000} & 0.260  & &3 & 0.106 & & 0.187 & \cellcolor[gray]{0.85}{0.210} & 0.107 & \cellcolor[gray]{0.85}{1.000} & 0.567  & &3 & 0.598 & & 0.033 & \cellcolor[gray]{0.85}{0.684} & 0.597 & \cellcolor[gray]{0.85}{1.000} & \cellcolor[gray]{0.85}{1.000} \\
& TCO & 2 & 0.285 & & 0.034 & \cellcolor[gray]{0.85}{0.813} & 0.296 & \cellcolor[gray]{0.85}{1.000} & 0.750  & &4 & 0.109 & & 0.318 & \cellcolor[gray]{0.85}{0.159} & 0.109 & \cellcolor[gray]{0.85}{1.000} & 0.950  & &1 & 0.591 & & 0.104 & \cellcolor[gray]{0.85}{0.728} & 0.590 & \cellcolor[gray]{0.85}{1.000} & \cellcolor[gray]{0.85}{1.000} \\
& ARM & 2 & 0.484 & & 0.002 & \cellcolor[gray]{0.85}{0.875} & 0.400 & \cellcolor[gray]{0.85}{1.000} & 0.400  & &--- & --- & & --- & --- & --- & --- & ---  & &--- & --- & & --- & --- & --- & --- & --- \\
& TUD & 6 & 0.099 & & 0.003 & \cellcolor[gray]{0.85}{0.617} & 0.109 & \cellcolor[gray]{0.85}{1.000} & 0.083  & &2 & 0.067 & & 0.234 & \cellcolor[gray]{0.85}{0.118} & 0.066 & \cellcolor[gray]{0.85}{1.000} & 0.900  & &--- & --- & & --- & --- & --- & --- & --- \\
\midrule
\multicolumn{2}{l}{\textbf{Mean}}
& & & & 0.101 & \cellcolor[gray]{0.85}{\underline{0.681}} & 0.242 & \cellcolor[gray]{0.85}{\underline{1.000}} & 0.412
& & & & & 0.303 & \cellcolor[gray]{0.85}{\underline{0.223}} & 0.174 & \cellcolor[gray]{0.85}{\underline{1.000}} & 0.876
& & & & & 0.040 & \cellcolor[gray]{0.85}{\underline{0.809}} & 0.680 & \cellcolor[gray]{0.85}{\underline{1.000}} & 0.988 \\

% ===================== NON-KILLABLE =====================
\midrule
\parbox[t]{2mm}{\multirow{11}{*}{\rotatebox[origin=c]{90}{\textit{Initially Non-Killable}}}}
& TAN & 2 & 0.088 & & 0.004 & \cellcolor[gray]{0.85}{0.619} & 0.092 & \cellcolor[gray]{0.85}{1.000} & 0.100  & &--- & --- & & --- & --- & --- & --- & ---  & &--- & --- & & --- & --- & --- & --- & --- \\
& RAW & --- & --- & & --- & --- & --- & --- & ---  & &--- & --- & & --- & --- & --- & --- & ---  & &1 & 0.448 & & 0.541 & \cellcolor[gray]{0.85}{0.502} & 0.000 & \cellcolor[gray]{0.85}{0.600} & 0.000 \\
& ACH & 3 & 0.079 & & 0.159 & \cellcolor[gray]{0.85}{0.259} & 0.082 & \cellcolor[gray]{0.85}{1.000} & 0.133  & &--- & --- & & --- & --- & --- & --- & ---  & &--- & --- & & --- & --- & --- & --- & --- \\
& HBS & 1 & 0.066 & & 1.000 & \cellcolor[gray]{0.85}{0.066} & \cellcolor[gray]{0.85}{0.066} & \cellcolor[gray]{0.85}{0.000} & \cellcolor[gray]{0.85}{0.000}  & &--- & --- & & --- & --- & --- & --- & ---  & &--- & --- & & --- & --- & --- & --- & --- \\
& HNE & --- & --- & & --- & --- & --- & --- & ---  & &5 & 0.038 & & 1.000 & \cellcolor[gray]{0.85}{0.038} & \cellcolor[gray]{0.85}{0.038} & \cellcolor[gray]{0.85}{0.000} & \cellcolor[gray]{0.85}{0.000}  & &--- & --- & & --- & --- & --- & --- & --- \\
& TCL & 4 & 0.088 & & 0.359 & \cellcolor[gray]{0.85}{0.243} & 0.090 & \cellcolor[gray]{0.85}{0.750} & 0.000  & &3 & 0.048 & & 1.000 & \cellcolor[gray]{0.85}{0.048} & \cellcolor[gray]{0.85}{0.048} & \cellcolor[gray]{0.85}{0.000} & \cellcolor[gray]{0.85}{0.000}  & &--- & --- & & --- & --- & --- & --- & --- \\
& HLR & --- & --- & & --- & --- & --- & --- & ---  & &2 & 0.046 & & 1.000 & \cellcolor[gray]{0.85}{0.046} & \cellcolor[gray]{0.85}{0.046} & \cellcolor[gray]{0.85}{0.000} & \cellcolor[gray]{0.85}{0.000}  & &--- & --- & & --- & --- & --- & --- & --- \\
& LCH & 1 & 0.068 & & 1.000 & \cellcolor[gray]{0.85}{0.068} & \cellcolor[gray]{0.85}{0.068} & \cellcolor[gray]{0.85}{0.000} & \cellcolor[gray]{0.85}{0.000}  & &--- & --- & & --- & --- & --- & --- & ---  & &--- & --- & & --- & --- & --- & --- & --- \\
& OCH & 3 & 0.086 & & 0.004 & \cellcolor[gray]{0.85}{0.535} & 0.086 & \cellcolor[gray]{0.85}{1.000} & 0.233  & &--- & --- & & --- & --- & --- & --- & ---  & &--- & --- & & --- & --- & --- & --- & --- \\
& CWI & 9 & 0.075 & & 0.228 & \cellcolor[gray]{0.85}{0.426} & 0.068 & \cellcolor[gray]{0.85}{0.778} & 0.056  & &--- & --- & & --- & --- & --- & --- & ---  & &--- & --- & & --- & --- & --- & --- & --- \\
& TRD & 1 & 0.084 & & 0.004 & \cellcolor[gray]{0.85}{0.750} & 0.100 & \cellcolor[gray]{0.85}{1.000} & 0.000  & &2 & 0.043 & & 1.000 & \cellcolor[gray]{0.85}{0.043} & \cellcolor[gray]{0.85}{0.043} & \cellcolor[gray]{0.85}{0.000} & \cellcolor[gray]{0.85}{0.000}  & &--- & --- & & --- & --- & --- & --- & --- \\
& TCO & 4 & 0.084 & & 0.515 & \cellcolor[gray]{0.85}{0.228} & 0.084 & \cellcolor[gray]{0.85}{0.500} & 0.000  & &2 & 0.056 & & 1.000 & \cellcolor[gray]{0.85}{0.056} & \cellcolor[gray]{0.85}{0.056} & \cellcolor[gray]{0.85}{0.000} & \cellcolor[gray]{0.85}{0.000}  & &1 & 0.208 & & 0.053 & \cellcolor[gray]{0.85}{0.364} & 0.000 & \cellcolor[gray]{0.85}{1.000} & 0.000 \\
\midrule
\multicolumn{2}{l}{\textbf{Mean}} 
& & & & 0.364 & \cellcolor[gray]{0.85}{\underline{0.355}} & 0.082 & \cellcolor[gray]{0.85}{\underline{0.670}} & 0.058 
& & & & & 1.000 & \cellcolor[gray]{0.85}{0.046} & \cellcolor[gray]{0.85}{0.046} & \cellcolor[gray]{0.85}{0.000} & \cellcolor[gray]{0.85}{0.000} 
& & & & & 0.296 & \cellcolor[gray]{0.85}{\underline{0.433}} & 0.000 & \cellcolor[gray]{0.85}{\underline{0.800}} & 0.000 \\
\bottomrule
\end{tabular}}
\end{table*}

The results of the mutation analysis for MN, CF and SR can be found in Table~\ref{Pipeline2ResultsA}, while Table~\ref{Pipeline2ResultsB} shows the results obtained for RE, UD and UE.
For each MO, we report the number of mutant configurations (column `\textit{\#}') along with the average percentage of test set inputs that contribute to killing mutants across configurations (column `\textit{$t_c$}').

Results reported for the $\rho_{\bm{k}}$ metric show that across subjects  on average 9.25\% of the test data is enough to kill a mutant. This allows us to pinpoint a specific region of the input space that can quickly lead to fault detection. This value goes to as low as 0.1\% for the RAW operator and as high as 99.5\% for TAN operator of the RE subject. For the UD dataset, the average $\rho_k$ across all MOs for killable mutants reaches 30.3\%, which is the highest across all subjects. 
    
The average proportion of contributors ($\rho_{\bm{c}}$) in the killing aggregation reaches 51.7\%. The smallest proportion is observed for UD (22.3\%), while the highest is observed for UE  (80.9\%). Across all the subjects and MOs, in 59\% of the subject/MO pairs, the proportion of contributors is 50\% or more, in 25\% it is at least 75\%, and in 11\%, the percentage ranges between 90\% and 100\%. This further supports the claim that the obtained killing aggregations are \emph{fault-discriminative}: % failure-discriminative
\topomap's killing aggregations are very small compared to their respective parent test set and, at the same time, the percentage of contributors within them is high enough to kill the mutants.
Furthermore, across the subjects, the average $\rho_{\bm{c}}$ of the aggregations produced with \topomap is always greater than the one measured on the random selection. In particular, in the case of MN, the concentration of contributors is three times bigger in the \topomap aggregations than in the random ones. Similar results are obtained for  RE and SR.

Column $\rho^{(R)}_{d}$ in Table~\ref{Pipeline2ResultsA} and Table~\ref{Pipeline2ResultsB} reports the killing probability of  a set of randomly selected inputs having the same size as the killing aggregation identified by our approach. It is worth mentioning that the probability of killing for our approach is 100\%, as by construction we aggregate regions in the input space that are together capable of killing a mutant.
The comparison shows that the increase in killing probability is approximately 58\%\textit{pt} for the RE and MN subjects, over 50\%\textit{pt} for the SR subject, and 26\%\textit{pt} and 12\%\textit{pt} for the CF and UD subjects, respectively. For the UE dataset, the difference in probability is around 1\%\textit{pt}, which is due to the mutants of this subject being killable by a diverse set of several unrelated inputs, so that a random selection is able to achieve good performance. The average gain across subjects when using our approach against random input selection is 35\%\textit{pt}.

Furthermore, in Table~\ref{Pipeline2ResultsA} and Table~\ref{Pipeline2ResultsB} we present the average killing probability results of \topomap vs random selection on the set of mutants not killed by the original test set (see the \emph{Initially Non-Killable} rows in the tables).
For UD, neither \topomap nor random selection could kill any additional mutant, confirming the hypothesis that these mutants are likely equivalent (i.e., non-distinguishable from the original program)~\cite{10.1145/3460319.3464825}.
For the other subjects, however, the situation is different. For the \emph{Initially Non-Killable} mutants of MN, \topomap achieves a killing probability of 93\%, representing an improvement of 89\%\textit{pt} over the random approach. For the remaining subjects (except for UD), the killing probability ranges from 67\% to 84\%, with an average improvement of 70\%\textit{pt} over the random selection. On average across all subjects except UD, the improvement is 74\%pt.

Although white-box approaches are not directly comparable with \topomap, which is instead completely black-box and hence is more widely applicable, we conducted an experiment, limited to MN, to understand the margin that white-box approaches may have over black-box ones. On MN we computed the concentration of contributors $\rho_c$ and the probability of killing $\rho_d$ achieved by DeepGini when selecting the same proportion $\rho_k$ of inputs as the one considered by \topomap. On killable mutants, DeepGini achieves $\rho_c =$ 45\% and $\rho_d =$ 70\%, while on initially non-killable mutants it achieves $\rho_c =$ 17\% and $\rho_d =$ 8\%. We cannot compare the two approaches on $\rho_d$, as the experiment with \topomap was designed to achieve maximum $\rho_d$ (which is confirmed by the substantial margin of \topomap over DeepGini). On $\rho_c$, DeepGini is superior to \topomap by 19.8\%pt on killable mutants and by 12.2\%pt on initially non killable ones, with an overall average margin of 16\%pt over \topomap. This confirms that as expected white-box methods do have a margin over black box ones, but the difference between them is relatively small and supports usage of black box methods like \topomap in  contexts in which white-box access is problematic or impossible.

\begin{custombox}[Answer to RQ3]
Our results show that \topomap is highly effective in locating input space regions with high density of inputs sharing features that expose faults in DNN models. The average advantage across subjects of \topomap compared to a random sampling is 35\%\textit{pt} on killable mutants and 74\%\textit{pt} on initially non-killable ones, using on average just 9\% of the input data.
\end{custombox}

\textbf{RQ4. Comparison with DeepHyperion:} \emph{How does the map produced by \topomap differ from that produced by the state-of-the-art technique DeepHyperion?}

For the MN subject, \topomap produces a topographical map consisting of 90 clusters. To enable a per-label comparison with DeepHyperion, we construct a label-specific sub-map by locating all the clusters that contain inputs of that label. We then bin the feature values of DeepHyperion to generate a DeepHyperion map ensuring that the number of populated cells is as close as possible to the number of clusters in the label-specific \topomap. %In this way, we obtain that \topomap identifies on average 395 sub-clusters, across 10 experimental runs, ranging from 391 to 400. On the other hand, DeepHyperion identifies on average 378 cells, ranging from 367 to 400, again across 10 empirical runs. \nargiz{I think it is better to report the average number of clusters/cells in per-label maps.}\gianmarco{Ok, I'll do it. thanks}
In this way, we obtain that \topomap identifies on average 39.50 sub-clusters per label, across 10 experimental runs. On the other hand, DeepHyperion identifies on average 37.82 cells per label, again across 10 empirical runs.
% 344
By comparing the maps generated by \topomap and DeepHyperion, we get a similarity score of 0.605 $\pm$ 0.006 (averaged across the 10 empirical runs), which implies a low similarity, hence complementarity, between the two maps.

Table~\ref{Pipeline2ResultsDeepHyperion} shows the comparison of the metrics computed after evaluating both \topomap and DeepHyperion on mutation killing. 
In both \emph{killable} and \emph{initially non-killable} mutants scenarios, \topomap generally outperforms DeepHyperion in terms of proportion of test inputs in aggregation, needing fewer inputs than the baseline to statistically significantly kill the mutants, as well as for the density of killing inputs in the killing aggregation made by the two maps: on average, aggregations gathered by \topomap are 8\%\textit{pt} denser in contributors than DeepHyperion. This occurs in almost all cases except for the ACH, TCL, HDB and TCO mutation operators. 

Likewise, for mutants that are not statistically significantly killed by the test set alone, we see that \topomap achieves higher density of contributors (89\% vs 77\%). However, on average DeepHyperion requires slightly fewer inputs to make up a killing aggregation (mean $\rho_k = $ $5.2\times 10^{-4}$ vs $5.5\times 10^{-4}$). At the same time, the generation of \topomap maps is completely automated and does not require intervention from human assessors, while DeepHyperion requires manual definition and assessment of discriminative features.

Besides that, it is interesting to notice the lack of symmetry across MOs in the metric $\rho_k$ between the two approaches. In fact, it is not necessarily true that the MOs requiring more inputs to be killed when using \topomap require a proportionally similar amount of inputs with DeepHyperion, i.e., there is no correspondence between the two approaches in terms of ``most demanding" MOs.

As for the impurity in the distribution of critical inputs across the regions of the map, Table~\ref{UncertaintyComparisonDeepHyperion} shows that on different model architectures trained for the same problem, both \topomap and DeepHyperion outperform each other in different scenarios, always achieving statistical significance (here computed through a paired t-test, where statistical significance is assessed for a $p$-value $<$ 0.05). Both techniques achieve a low impurity with all values being close to 0, especially for \topomap, with impurities ranging from 0.025 to 0.135, meaning that maps effectively separate critical inputs from non-critical ones.

\begin{table}[t]
\caption{Proportion $\rho_k$ of inputs included into the killing cluster aggregation for the MN dataset averaged over ten empirical runs. The density of contributing inputs $\rho_c$ quantifies the concentration of killing inputs in the considered cluster aggregations. Results are compared with the concentration of contributors $\rho_c$ and the probability $\rho_d$ of the mutant being killed by the aggregation obtained from the feature regions identified by DeepHyperion.}\label{Pipeline2ResultsDeepHyperion}%
\small
\centering
\begin{tabular}{@{}cccccccccccccccccccccccccccccccccccccccccccccccc@{}}
    \toprule
    % Table column names
    & \textbf{MO} & \textbf{\#} & $\bm{t_c}$ & & $\bm{\rho_\bm{k}}$ & $\bm{\rho^{(DH)}_\bm{k}}$ & $\bm{\rho_\bm{c}}$ & $\bm{\rho^{(DH)}_\bm{c}}$ & $\bm{\rho_d}$ & $\bm{\rho^{(DH)}_d}$ \\
    \midrule
    \parbox[t]{2mm}{\multirow{15}{*}{\rotatebox[origin=c]{90}{\textit{Killable}}}}
& TAN & 2 & 0.014 & & \cellcolor[gray]{0.85}{1.5e-04} & 3.5e-04 & \cellcolor[gray]{0.85}{0.983} & 0.800 & \cellcolor[gray]{0.85}{1.000} & \cellcolor[gray]{0.85}{1.000} \\
& RAW & 3 & 0.013 & & \cellcolor[gray]{0.85}{2.8e-04} & 1.0e-03 & \cellcolor[gray]{0.85}{0.954} & 0.634 & \cellcolor[gray]{0.85}{1.000} & \cellcolor[gray]{0.85}{1.000} \\
& ACH & 9 & 0.019 & & \cellcolor[gray]{0.85}{2.8e-04} & 3.5e-04 & 0.942 & \cellcolor[gray]{0.85}{0.966} & \cellcolor[gray]{0.85}{1.000} & \cellcolor[gray]{0.85}{1.000} \\
& HBS & 3 & 0.015 & & \cellcolor[gray]{0.85}{2.1e-04} & 5.4e-04 & \cellcolor[gray]{0.85}{0.957} & 0.872 & \cellcolor[gray]{0.85}{1.000} & \cellcolor[gray]{0.85}{1.000} \\
& HNE & 3 & 0.023 & & \cellcolor[gray]{0.85}{3.2e-04} & 9.1e-04 & \cellcolor[gray]{0.85}{0.884} & 0.847 & \cellcolor[gray]{0.85}{1.000} & \cellcolor[gray]{0.85}{1.000} \\
& TCL & 6 & 0.043 & & 5.0e-04 & \cellcolor[gray]{0.85}{3.5e-04} & 0.904 & \cellcolor[gray]{0.85}{0.947} & \cellcolor[gray]{0.85}{1.000} & \cellcolor[gray]{0.85}{1.000} \\
& HLR & 3 & 0.098 & & \cellcolor[gray]{0.85}{3.1e-04} & 5.2e-04 & \cellcolor[gray]{0.85}{0.983} & 0.767 & \cellcolor[gray]{0.85}{1.000} & \cellcolor[gray]{0.85}{1.000} \\
& LCH & 10 & 0.028 & & \cellcolor[gray]{0.85}{1.4e-04} & 2.8e-04 & \cellcolor[gray]{0.85}{0.997} & 0.990 & \cellcolor[gray]{0.85}{1.000} & \cellcolor[gray]{0.85}{1.000} \\
& OCH & 3 & 0.028 & & \cellcolor[gray]{0.85}{2.1e-04} & 3.7e-04 & \cellcolor[gray]{0.85}{1.000} & 0.864 & \cellcolor[gray]{0.85}{1.000} & \cellcolor[gray]{0.85}{1.000} \\
& CWI & 4 & 0.453 & & \cellcolor[gray]{0.85}{2.3e-04} & 2.5e-04 & \cellcolor[gray]{0.85}{0.969} & 0.925 & \cellcolor[gray]{0.85}{1.000} & \cellcolor[gray]{0.85}{1.000} \\
& TRD & 4 & 0.114 & & \cellcolor[gray]{0.85}{2.0e-04} & 5.3e-04 & \cellcolor[gray]{0.85}{0.978} & 0.882 & \cellcolor[gray]{0.85}{1.000} & \cellcolor[gray]{0.85}{1.000} \\
& HDB & 1 & 0.015 & & \cellcolor[gray]{0.85}{1.6e-04} & 3.3e-04 & 0.975 & \cellcolor[gray]{0.85}{0.983} & \cellcolor[gray]{0.85}{1.000} & \cellcolor[gray]{0.85}{1.000} \\
& TCO & 1 & 0.122 & & 3.8e-04 & \cellcolor[gray]{0.85}{2.0e-04} & \cellcolor[gray]{0.85}{1.000} & \cellcolor[gray]{0.85}{1.000} & \cellcolor[gray]{0.85}{1.000} & \cellcolor[gray]{0.85}{1.000} \\
& ARM & 4 & 0.260 & & \cellcolor[gray]{0.85}{2.8e-04} & 4.1e-04 & \cellcolor[gray]{0.85}{0.954} & 0.851 & \cellcolor[gray]{0.85}{1.000} & \cellcolor[gray]{0.85}{1.000} \\
& TUD & 6 & 0.037 & & 4.4e-04 & \cellcolor[gray]{0.85}{4.3e-04} & \cellcolor[gray]{0.85}{0.956} & 0.908 & \cellcolor[gray]{0.85}{1.000} & \cellcolor[gray]{0.85}{1.000} \\
    \midrule
    \multicolumn{2}{l}{\textbf{Mean}} & & & & \cellcolor[gray]{0.85}{\underline{2.7e-04}} & 4.5e-04 & \cellcolor[gray]{0.85}{\underline{0.962}} & 0.882 & \cellcolor[gray]{0.85}{1.000} & \cellcolor[gray]{0.85}{1.000}\\
    \midrule
    \parbox[t]{2mm}{\multirow{7}{*}{\rotatebox[origin=c]{90}{\textit{\footnotesize{Init. Non-Killable}}}}}
& TAN & 2 & 0.011 & & 5.4e-04 & \cellcolor[gray]{0.85}{5.2e-04} & \cellcolor[gray]{0.85}{0.834} & 0.595 & \cellcolor[gray]{0.85}{1.000} & \cellcolor[gray]{0.85}{1.000} \\
& HNE & 2 & 0.011 & & \cellcolor[gray]{0.85}{5.1e-04} & 7.3e-04 & \cellcolor[gray]{0.85}{0.924} & 0.638 & \cellcolor[gray]{0.85}{1.000} & \cellcolor[gray]{0.85}{1.000} \\
& TCL & 1 & 0.010 & & 5.1e-04 & \cellcolor[gray]{0.85}{2.0e-04} & 0.862 & \cellcolor[gray]{0.85}{1.000} & \cellcolor[gray]{0.85}{1.000} & \cellcolor[gray]{0.85}{1.000} \\
& HLR & 2 & 0.010 & & \cellcolor[gray]{0.85}{6.5e-04} & 6.8e-04 & \cellcolor[gray]{0.85}{0.837} & 0.585 & \cellcolor[gray]{0.85}{1.000} & \cellcolor[gray]{0.85}{1.000} \\
& TRD & 3 & 0.012 & & \cellcolor[gray]{0.85}{5.4e-04} & 7.0e-04 & \cellcolor[gray]{0.85}{0.893} & 0.709 & \cellcolor[gray]{0.85}{1.000} & \cellcolor[gray]{0.85}{1.000} \\
& TCO & 2 & 0.011 & & 5.3e-04 & \cellcolor[gray]{0.85}{5.0e-04} & \cellcolor[gray]{0.85}{0.894} & 0.875 & \cellcolor[gray]{0.85}{1.000} & \cellcolor[gray]{0.85}{1.000} \\
& TUD & 1 & 0.013 & & 5.8e-04 & \cellcolor[gray]{0.85}{3.0e-04} & 0.986 & \cellcolor[gray]{0.85}{1.000} & \cellcolor[gray]{0.85}{1.000} & \cellcolor[gray]{0.85}{1.000} \\
    \midrule
    \multicolumn{2}{l}{\textbf{Mean}} & & & & 5.5e-04 & \cellcolor[gray]{0.85}{\underline{5.2e-04}} & \cellcolor[gray]{0.85}{\underline{0.890}} & 0.772 & \cellcolor[gray]{0.85}{1.000} & \cellcolor[gray]{0.85}{1.000}\\
    \bottomrule
\end{tabular}
\end{table}

\begin{table}[t]
\caption{Gini impurity for each mapping technique indicating their ability of separating critical (high-uncertainty and misclassified) inputs from non-critical ones across different architectures. Bold values indicate statistical significance ($p$-value $< 0.05$).}\label{UncertaintyComparisonDeepHyperion}%
\small
\centering
\begin{tabular}{@{}lrcccccccccc@{}}
    \toprule
    & & & \multicolumn{2}{c}{\textbf{DeepGini}} & & \multicolumn{2}{c}{\textbf{Vanilla SM}} & & \multicolumn{2}{c}{\textbf{Misclassified}} \\
    \cmidrule{4-5}\cmidrule{7-8}\cmidrule{10-11}
    \textbf{Model} & & & \textbf{TM} & \textbf{DH} & & \textbf{TM} & \textbf{DH} & & \textbf{TM} & \textbf{DH}\\
    \midrule
    7-layer CNN & \cite{dartrisenkerasmnist}  & & \cellcolor[gray]{0.85}{\textbf{0.046}} & 0.049 & & \cellcolor[gray]{0.85}{\textbf{0.042}} & 0.058 & & \cellcolor[gray]{0.85}{\textbf{0.029}} & 0.039\\
    8-layer CNN & \cite{chollet2015keras}     & & 0.109 & \cellcolor[gray]{0.85}{\textbf{0.087}} & & 0.109 & \cellcolor[gray]{0.85}{\textbf{0.087}} & & 0.040 & \cellcolor[gray]{0.85}{\textbf{0.021}}\\
    19-layer CNN & \cite{kj7kunalmnistkeras}  & & 0.101 & \cellcolor[gray]{0.85}{\textbf{0.090}} & & 0.101 & \cellcolor[gray]{0.85}{\textbf{0.090}} & & 0.025 & \cellcolor[gray]{0.85}{\textbf{0.021}}\\
    LeNet1 & \cite{10.1145/3644388}      & & \cellcolor[gray]{0.85}{\textbf{0.027}} & 0.028 & & \cellcolor[gray]{0.85}{\textbf{0.039}} & 0.045 & & \cellcolor[gray]{0.85}{\textbf{0.135}} & 0.184\\
    LeNet5 & \cite{10.1145/3644388}      & & \cellcolor[gray]{0.85}{\textbf{0.052}} & 0.059 & & \cellcolor[gray]{0.85}{\textbf{0.062}} & 0.067 & & \cellcolor[gray]{0.85}{\textbf{0.131}} & 0.192\\
    VGG11 & \cite{vg11mnist}             & & 0.115 & \cellcolor[gray]{0.85}{\textbf{0.093}} & & 0.114 & \cellcolor[gray]{0.85}{\textbf{0.093}} & & 0.102 & \cellcolor[gray]{0.85}{\textbf{0.066}}\\
    \bottomrule
\end{tabular}
\end{table}  

\begin{custombox}[Answer to RQ4] Empirical results show that \topomap produces a map that is semantically different from the one produced by DeepHyperion, and outperforms it on the MNIST subject in terms of contributors density by 8\%\textit{pt} in killable mutants and by roughly 12\%\textit{pt} on initially non-killable mutants, in both cases with a smaller aggregation of test inputs.
\end{custombox}

\textbf{RQ5. Killing Propagation:} \emph{Do different aggregations, which kill different MOs and mutation configurations, repeatedly include the same clusters?}
%We investigate whether there exist regions in the topographical map that consistently participate in the fault exposure. 

Results presented in Table~\ref{ResultsKillingPropagationRhoA} show that the killing capability of the first most selected cluster (row `\emph{First}') reaches 67\% of mutants for MN, 65\% for UD, 57\% for UE, and drops to 45\%, 39\% and 27\% for SR, RE, and CF, respectively. The highest participation in killing aggregations is observed for the second (35\%) and third MSC (24\%) for MN, the lowest second MSC (19\%) for CF, and third MSC (5\%) for UE.
%\new{From Table \ref{ResultsKillingPropagationPercRhoA} we can see that for all subjects there are 2\% (MN) to 40\% (UD) of clusters that participate in 25\% or more killing aggregations. For RE, the percentage of clusters that participate in killing aggregations for at least 15\% mutants increases to more than 10\%, while it remains the same for UD, and increases to 4\%-8\% for the remaining subjects.} \nargiz{the last sentence is not very clear}

    \begin{table}[!t]
    %\parbox{0.5\linewidth}{
        \caption{Proportion $\rho_a$ of aggregations that contain the 1st, 2nd, 3rd most selected cluster (MSC) across the killable mutants per subject.}\label{ResultsKillingPropagationRhoA}
        \small
        \centering    
    \begin{tabular}{@{}lrrrrrr@{}}
    \toprule
    \textbf{MSC}    & \multicolumn{1}{c}{\textbf{MN}} 
                    & \multicolumn{1}{c}{\textbf{CF}} 
                    & \multicolumn{1}{c}{\textbf{SR}} 
                    & \multicolumn{1}{c}{\textbf{RE}} 
                    & \multicolumn{1}{c}{\textbf{UD}}
                    & \multicolumn{1}{c}{\textbf{UE}}\\
    \midrule
    First  & 67.74\% & 27.17\% & 45.23\% & 39.13\% & 65.38\% & 57.21\%\\
    Second & 35.16\% & 19.78\% & 31.82\% & 33.33\% & 34.62\% & 26.72\%\\
    Third  & 24.52\% & 16.30\% & 17.95\% & 21.74\% & 11.54\% & 5.08\%\\
    \bottomrule
    \end{tabular} %}
%     \hfill
%     \parbox{.49\linewidth}{
%        \caption{Proportion of clusters whose $\rho_a$ is greater  than respectively 25\%, 15\% and 5\%, across the killable mutants per subject.}\label{ResultsKillingPropagationPercRhoA}
%         \resizebox*{\linewidth}{!}{\begin{tabular}{@{}lcccccc@{}}
%     \toprule
%     $\bm{\rho_a}$ & \textbf{MN} & \textbf{CF} & \textbf{SR} & \textbf{RE} & \textbf{UD} & \textbf{UE}\\
%     \midrule
% %    $> 50\%$ & 0.011 & 0.000 & 0.000 & 0.000 & 0.200 & 0.037\\
%     $> 25\%$ & 0.022 & 0.025 & 0.057 & 0.024 & 0.400 & 0.074\\
%     $> 15\%$ & 0.044 & 0.075 & 0.086 & 0.108 & 0.400 & 0.074\\
%     %$> 10\%$ & 0.067 & 0.150 & 0.114 & 0.169 & 0.600 & 0.074\\
%     $> \phantom{0}5\%$ & 0.122 & 0.350 & 0.143 & 0.277 & 0.600 & 0.111\\
%     \bottomrule
%     \end{tabular}}}
    \end{table}

Additionally, we analysed whether different configurations of the same MO tend to be killed by the same aggregations. In Table~\ref{tab:kill_analysis}, we present the results at both the mutation operator and the subject levels.  For each mutation operator in column `\emph{MO}', columns with subject names show the number of configurations this operator has for each subject. Column `\emph{Avg\#Cfgs}' shows the average number of configurations across all subjects for each operator. Column `\emph{KillHalf}' shows the probability that the killing aggregation selected by \topomap kills at least half of the MO's configurations, and column `\emph{KillFull}' shows the probability that the aggregation kills all the configurations of the MO. At the bottom of the table, row `\emph{\#MOs}' shows the number of MOs applied per subject and rows `\emph{KillHalf}' and `\emph{KillFull}' show the average probabilities per subject calculated across all applicable operators.

Results indicate that `\emph{KillHalf}' probability is at least 80\% for 71\% of the operators. The highest probability of 100\% is observed for the AAL, RCP, HDB, TCO, ARM operators, and the lowest probability is reached by the RAW operator (55\%).  Moreover, this probability reaches 100\% on at least one subject for 15 operators out of 17 (AAL, TAN, ACH, HNE, HBS, RCP, HDB, TCL, HLR, LCH, OCH, CWI, TCO, ARM, TUD).  When it comes to `\emph{KillFull}' probability, the results range from 0\% (AAL, ARM) to 100\% (HDB) with an average of 34\%. %However, the HDB operator is not the most interesting case, since it on average has only 1 configuration per subject. For other operators reaching high \textit{`KillFull'} probability ($> 50\%$), the average number of configurations is 3.2.
%The majority of these operators are aimed at mutating training data or training hyperparameters, which implies that their parameters have continuous values. This peculiarity allows generating mutants that exhibit more similar behaviour than categorical mutations (such as `change loss function'). \nargiz{I do not think this is the case anymore}

At the subject level,  results show that the probability that  \topomap's input aggregation kills no less than half of the MO configurations (`\emph{KillHalf}' row) is higher than 60\% for all subjects. In particular, the median probability is 92.5\% with the lowest achieved on the RE subject (63\%) and the highest on the UD and SR subjects (100\%). When it comes to the probability of killing all the configurations of an MO by the same killing aggregation (`\emph{KillFull}' row), the lowest results are observed for the RE (7\%) and UE (22\%) subjects. For the remaining subjects, it ranges between 30\% (CF) and 75\% (UD). These results indicate that our approach can successfully detect the intrinsic characteristics of the faults and identify discriminative killing input regions in the topographical map.

\begin{table}[t]
    \caption{Probability that the input cluster aggregation selected by \topomap kills at least half (\emph{KillHalf}) or all (\emph{KillFull}) of the MO configurations across different MO categories and subjects.}
    \label{tab:kill_analysis}
    \small
    \centering
    \setlength{\tabcolsep}{4pt}
    \begin{tabular}{@{}lccccccrrr@{}}
    \toprule
    & \multicolumn{6}{c}{\textbf{\#Configurations}}\\
    \cmidrule{2-7}
    \textbf{MO} & \textbf{MN} & \textbf{CF} & \textbf{SR} & \textbf{RE} & \textbf{UD} & \textbf{UE} & \textbf{Avg\#Cfgs} & \textbf{KillHalf} & \textbf{KillFull}\\
    \midrule
        AAL & --- & ---  & --- & --- & --- & 7 & 7.00    & 100\% & 0\%\\
        TAN & 2   & 4    & --- & 1   &---  & 3 & 2.50 & 78\%  & 50\%\\
        RAW &3    &12    &---  & 6   &---& 11 & 8.00& 55\%& 12\% \\
        ACH &9    & 5    & 1   & 14&---& 5& 6.80& 72\%& 36\% \\
        HBS &3    & 1    &---  & 3&---& 2 & 2.25& 95\%& 62\% \\
        RCP &---  &---   & 4   &---&---&---& 4.00& 100\%& 60\% \\
        HNE & 3   & 3    & 4   & 2& 2& 4 & 3.00& 98\%& 28\% \\
        TCL & 6   & 4    & 6   & 3& 1& 4 & 4.00& 93\%& 23\% \\
        HLR & 3   &---   &---  & 8& 1& 3 & 3.75& 95\%& 40\% \\
        LCH & 10  &---   &12   & 9& 10& 9 & 10.00 & 80\%& 28\% \\
        OCH & 3   & 4    & 4   & 3& 3& 2 & 3.17& 85\%& 22\% \\
        CWI & 4   & 3    & 1   & 5&---& 7 & 4.00& 88\%& 24\% \\
        TRD & 4   & 5    & 6   & 5& 3& 3 & 4.33& 72\%& 3\% \\
        HDB & 1   & 1    &---  &---&---&--- &  1.00& 100\%& 100\% \\
        TCO & 1   & 1    & 6   & 2& 4& 1 & 2.50& 100\%& 67\% \\
        ARM & 4   &---   &---  & 2&---&--- & 3.00& 100\%& 0\% \\
        TUD & 6   & 3    &---  & 6& 2&--- & 4.25& 60\%& 25\%\\
    \midrule
    \textbf{\#MOs} & 15 & 12 & 9 & 14 & 8 & 13\\
    \textbf{KillHalf} & \multicolumn{1}{r}{97\%} & \multicolumn{1}{r}{70\%} & \multicolumn{1}{r}{100\%} & \multicolumn{1}{r}{63\%} & \multicolumn{1}{r}{100\%} & \multicolumn{1}{r}{88\%} \\
    \textbf{KillFull} & \multicolumn{1}{r}{39\%} & \multicolumn{1}{r}{30\%} & \multicolumn{1}{r}{42\%} & \multicolumn{1}{r}{7\%} & \multicolumn{1}{r}{75\%} & \multicolumn{1}{r}{22\%} \\
    \bottomrule
    \end{tabular}
\end{table}

%To assess whether across different scenarios there exist some cluster that are recurrently included into the killing aggregation, we rely on the percentage $\rho_a$. From Table~\ref{ResultsKillingPropagationRhoA}, we observe that in all data scenario there is at least one cluster recurrently appearing into the aggregations assembled to kill different mutants. In particular, we observe that in the cases of MN and RE, the most selected cluster is included into more than half of the killing aggregations. As a general assessment, we infer from Table \ref{ResultsKillingPropagationPercRhoA} that, for each data scenario, there is a subset of clusters that is consistently included in the aggregation killing at least 10\% of the mutants.

\begin{custombox}[Answer to RQ5] 
Results confirm our hypothesis that there exist regions in the topographical map that consistently contribute to fault exposure when different faults are injected by mutation analysis.
%For each analyzed data scenario, there exists at least one cluster grouping together instances able to kill different mutants. 
\end{custombox}

% \subsection{Killing Aggregations}
\subsection{Visualisation}
\begin{figure*}[t]
    \centering
    \includegraphics[width=\textwidth]{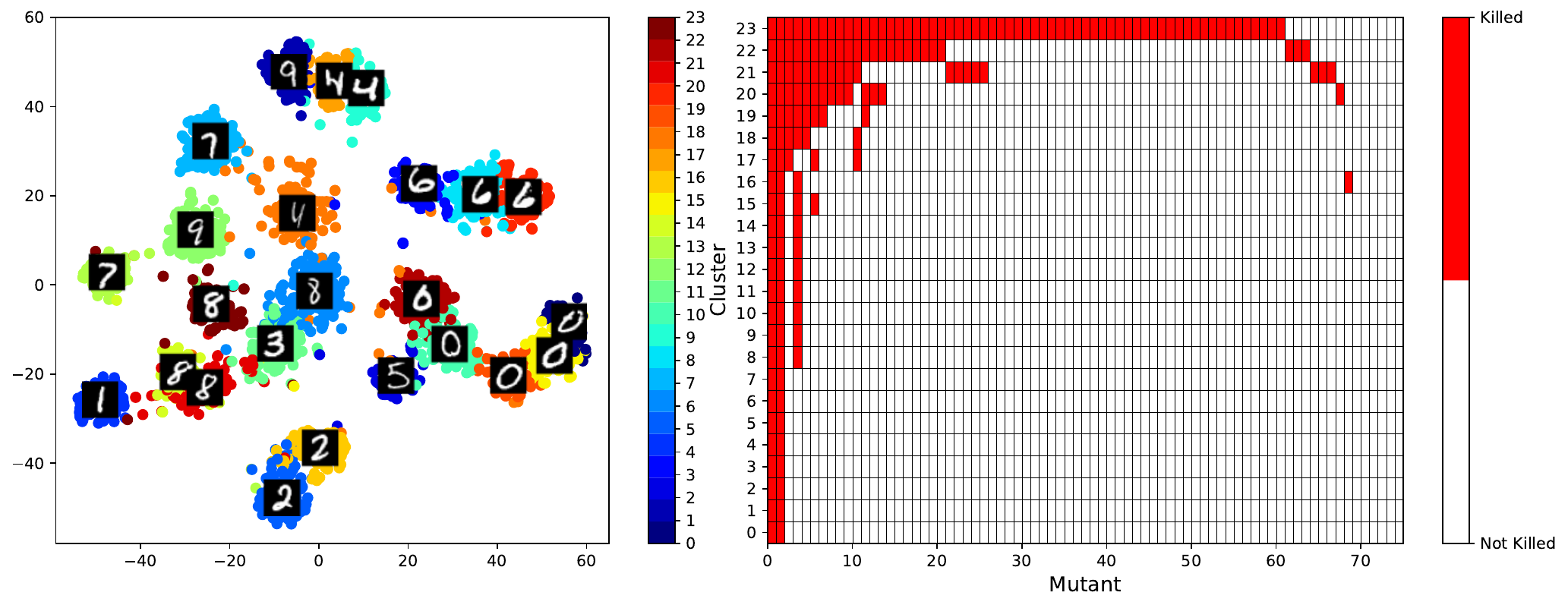}
    \caption{Topographical map computed on the MNIST dataset (SVD, K-means, 90)}
    \label{fig:topomap_mnist}
\end{figure*}

\begin{figure}[t]
    \centering
    \includegraphics[width=0.5\linewidth]{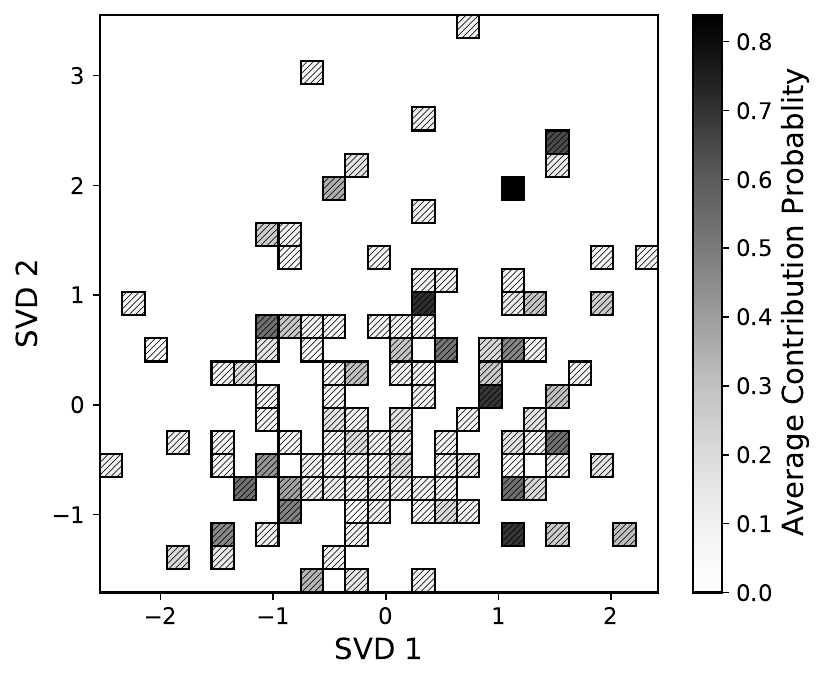}
    \caption{Average contribution probability within the most selected cluster on the MNIST dataset (SVD, K-means, 90)}
    \label{fig:topomap_mnist_msc}
\end{figure}

To better understand the structure and the properties of generated clusters, we provide a visual representation of a topographical map. Visualisation helps reveal how clusters relate to each other in the embedding space and highlights patterns that may not be evident from numerical results alone.

Fig.~\ref{fig:topomap_mnist} presents a selection of clusters from the topographical map generated by \topomap for the MN dataset, using SVD as the embedding method and K-means with 90 clusters as the clustering algorithm. The clusters are ranked by the ratio of mutants killed to the total number of mutants considered, and only those with a ratio greater than 0.05 are shown. This results in 24 clusters out of 90, which together kill 92\% of all available mutants. To improve interpretability, in Fig.~\ref{fig:topomap_mnist} we display the medoid of each cluster. The figure illustrates how groups of similar digits are positioned close to each other in the space and how stylistic features of the digits remain consistent across nearby clusters.

Fig.~\ref{fig:topomap_mnist_msc} shows the spread of contributors within the most selected cluster in the map of Fig.~\ref{fig:topomap_mnist}. For visualisation purposes, inputs have been binned according to the first two embedding components, with bordered and barred cells representing non-empty bins, while darker cells indicate inputs with a higher contribution probability, meaning that such inputs are responsible for killing multiple mutants. This representation shows that, although most sub-regions within the cluster contribute to mutant killing, certain areas exhibit a higher density of fault-revealing inputs, indicating that specific feature combinations are particularly promising from a testing perspective.

\begin{figure*}[t]
    \centering
    \subfloat[Digit ``5", MNIST (SVD, K-means, 90)]
    {\includegraphics[width=0.49\linewidth]{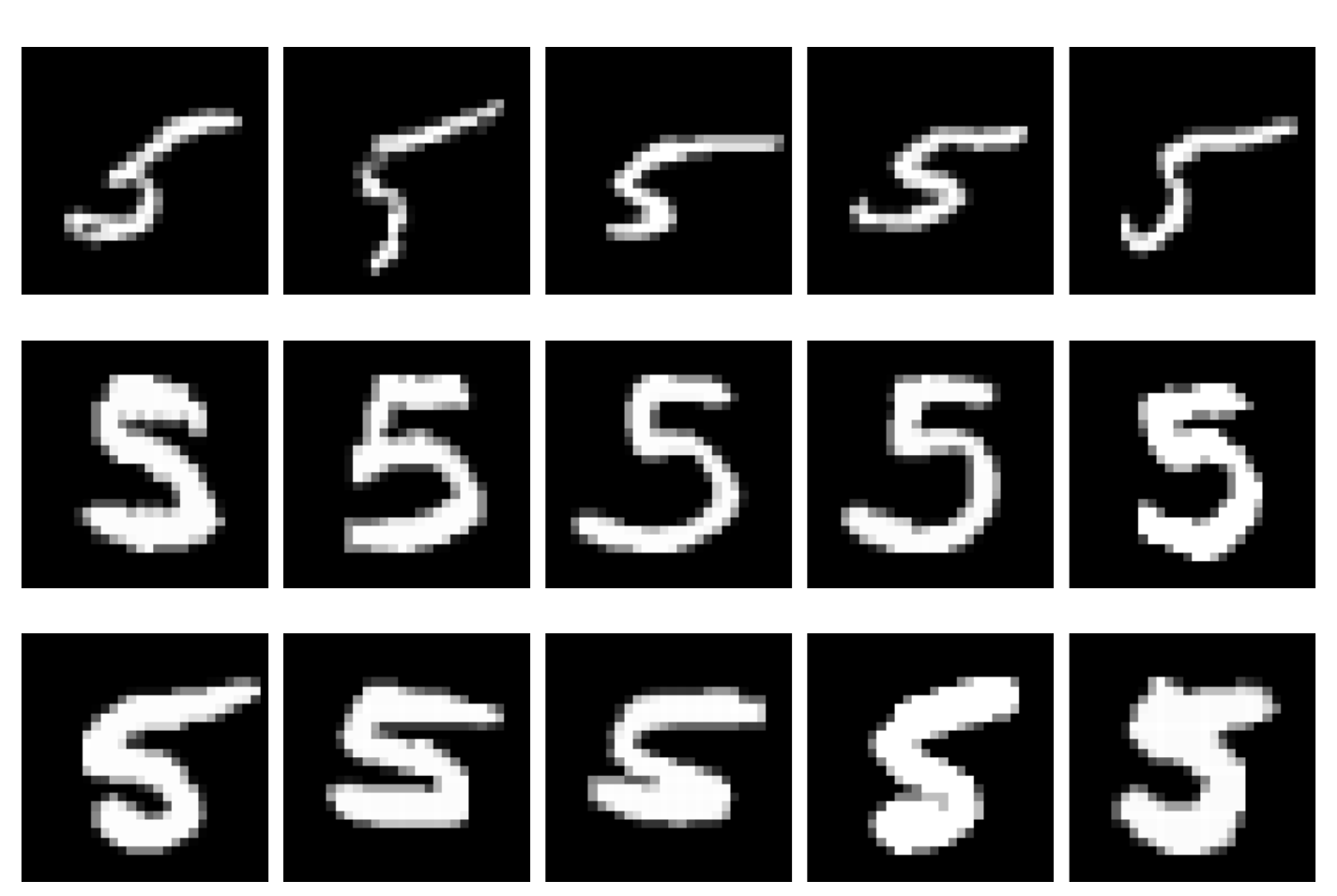}}
    \hfill
    \subfloat[Class ``horse", CIFAR-10 (PCA, K-means, 40)]
    {\includegraphics[width=0.49\linewidth]{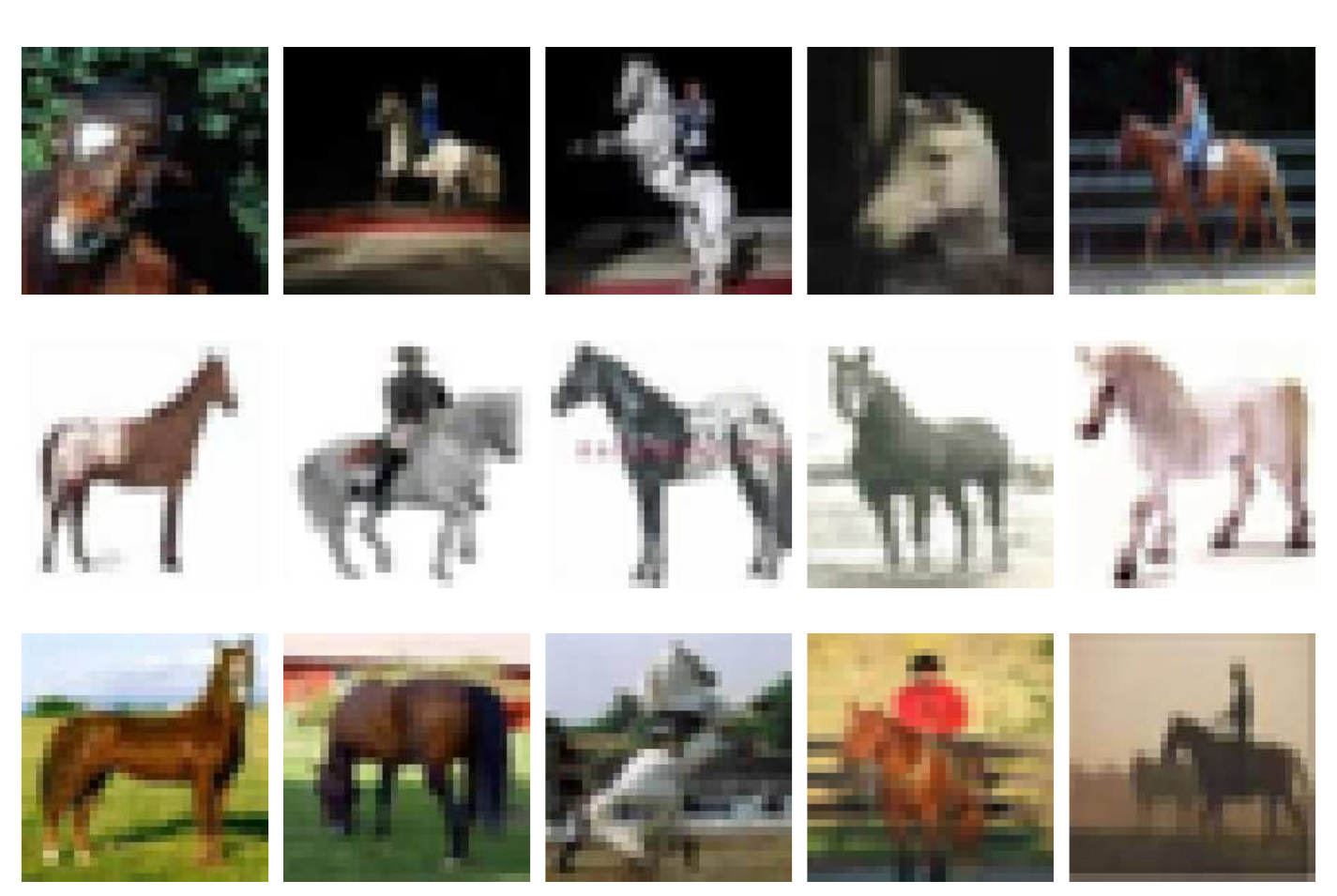}}
    \caption{Cohesion and separability of clusters in MNIST and CIFAR-10}
    \label{fig:cluster_visualisation_mnist}
\end{figure*}

Fig.~\ref{fig:cluster_visualisation_mnist} shows examples of \emph{recognisability} and \emph{separability} across three clusters selected from the topographical maps of the MN and CF subjects. On the left side of the panel, the samples of digit 5 within the three clusters (one per row) exhibit strong internal feature cohesion, characterised by light-stroke skewed digits (first row), thick-stroke expanded digits (second row), and thick-stroke compact digits (third row). Similarly, on the right side of the panel, the samples of horse images show consistent visual patterns in terms of their overall colour texture, including dark-background images (first row), white or transparent-background images (second row), and light open-air background images (third row). These patterns suggest that the clustering is mainly driven by image appearance and pixel-level characteristics, which is expected given that the selected embedding method (SVD/PCA) primarily captures low-level visual features. At the same time, the three clusters remain clearly distinguishable from one another, demonstrating their external separability.

To visualise a complete topographical map, we adopted a graph-based representation. Speci\-fi\-cally, each map is converted to a fully connected, weighted, undirected graph. In this graph, the vertices represent clusters and the edges encode the connections between clusters. The weight of each edge corresponds to the Euclidean distance between the centroids of the connected clusters. After computing all distances, we construct a minimum spanning tree of the graph and remove all edges with distances greater than the maximum distance in the tree. This pruning step ensures that the visualisation remains clear and interpretable.

\begin{figure*}[t]
    \centering
    \subfloat[MNIST (SVD, K-means, 90)%
        \label{fig:topomap_mnist_graph}]
    {\includegraphics[width=0.49\linewidth]{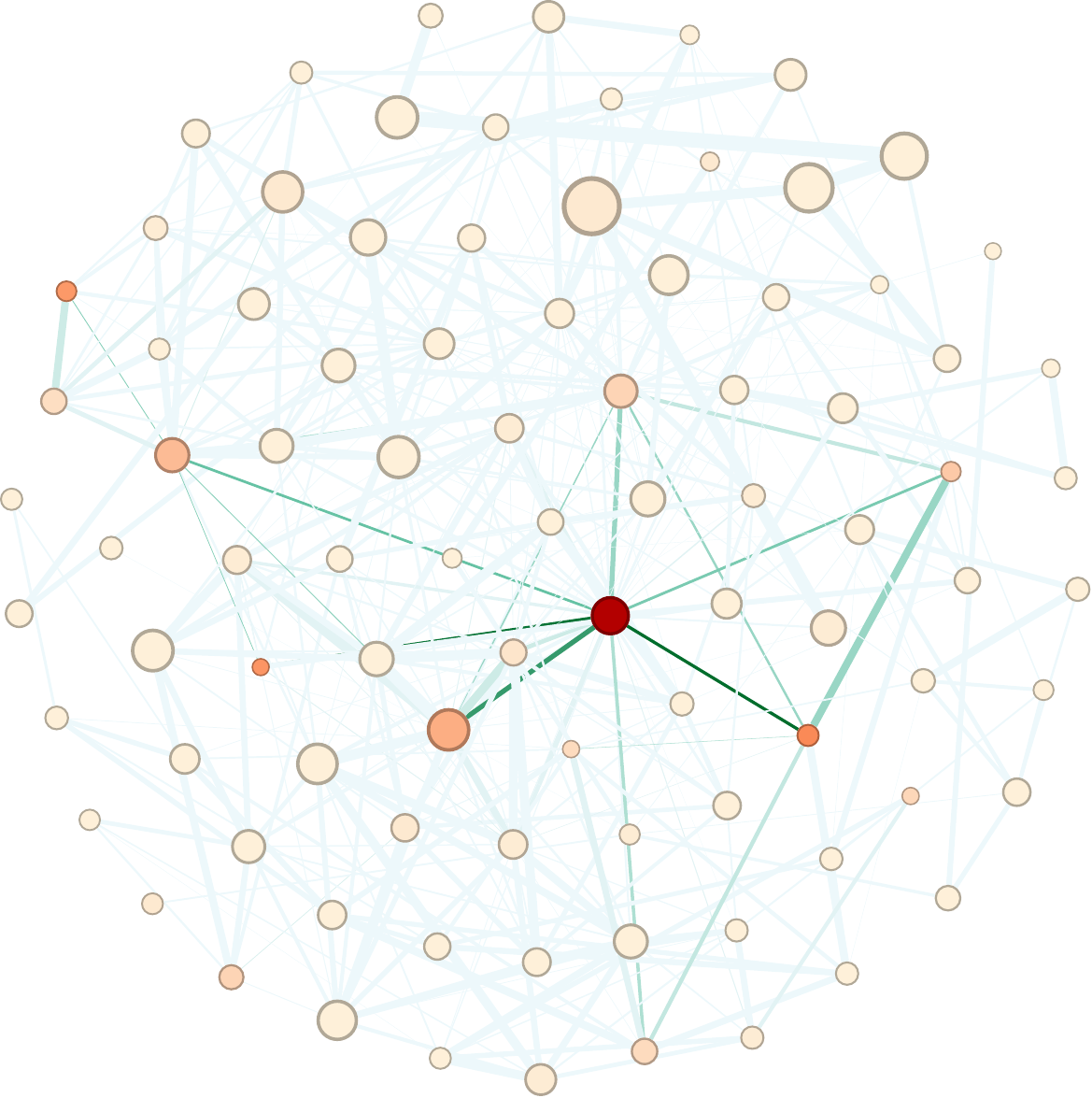}}
    \hfill
    \subfloat[Speaker Recognition (t-SNE, K-means, 40)%
        \label{fig:topomap_audio_graph}]
    {\includegraphics[width=0.49\linewidth]{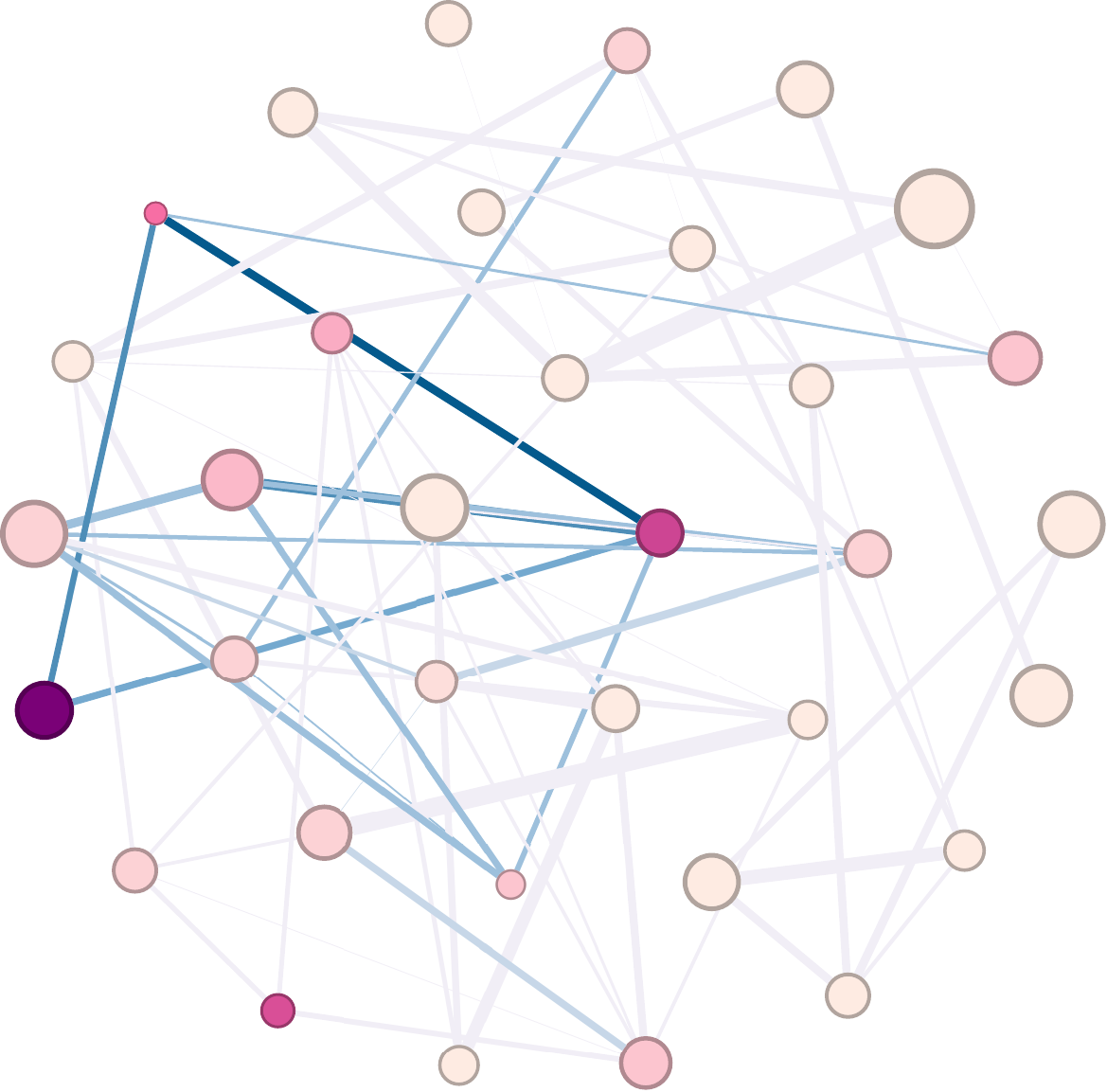}}
    \caption{Graph-based visualisation of the topographical maps for MNIST and Speaker Recognition subjects}
    \label{fig:topomap_graphs}
\end{figure*}

For rendering, we use Gephi~\cite{ICWSM09154}. Fig.~\ref{fig:topomap_graphs} shows graph representations of the maps generated by \topomap for the MN and SR datasets. The size of each vertex reflects the number of inputs falling into a cluster, while the colour intensity indicates the mutation-killing capability of that cluster. Edge thickness and colour intensity correspond to the distances between clusters. The visualisation highlights strong spatial relationships, showing that clusters with higher fault-revealing power tend to be closely connected.

\subsection{Threats to Validity}

\noindent \hspace{\parindent} \textbf{Construct Validity.} The choice of evaluation metrics could threaten our conclusions. To mitigate this risk, we did not limit ourselves to the most popular metrics such as the Silhouette score, but investigated additional measures that better capture the discriminative quality of clusters. For the mutation analysis, we refer to the statistical definition~\cite{9159089} of mutation killing. Different definitions could lead to different results.

\noindent \hspace{\parindent} \textbf{Internal Validity.} The main internal threat comes from the selection of dimensionality reduction techniques and clustering algorithms, as well as the tuning of their hyperparameters, which could affect the results. For this aspect, both the number of embedding dimensions and number of clusters are crucial, and we relied on established conventions for the former, and to an \emph{ad hoc} accuracy-based routine for the latter. Therefore, we experimented with the most frequently used approaches and performed a sophisticated evaluation to choose the best performing configuration for each subject. Furthermore, the empirical pipeline includes some steps (embedding with UMAP, clustering algorithms with K-means, training of a DNN-based assessor) that are non-deterministic, as they involve randomness at some point of their execution. We addressed this issue by running the experiments multiple times.

\noindent \hspace{\parindent} \textbf{External Validity.} We have evaluated \topomap on 6 subjects, and therefore there is no guarantee of generalisation. We made a careful choice to represent different tasks (classification and regression) from different application domains (image, speech, text). Although the number of empirical runs allows us to account for randomness and ensure the reliability of the results, clustering algorithms---as well as any learning method---are highly dependent on hyperparameters and the structure of the input data. Moreover, while the experimental design involves a total of 388 mutants grouped into multiple mutation operators, they do not account for all virtually possible DL faults and for all possible DNN architectures. Finally, although the empirical approach generalises well for the considered data scenarios, it cannot be ensured without further empirical evidence that such approach generalises for any application scenario.
\section{Discussion}\label{sec:discussion}
The analysis of the empirical results shows the existence of a link between fault-revealing capabilities of inputs and their location in the feature space. This aligns with previous work testing DNNs leveraging the exploration of the input space~\cite{10.1145/3460319.3464811}. 

This positions \topomap as a support framework to search-based DL testing, allowing for agnosticism w.r.t. the model under test, similarly to what has been achieved by tools like DeepHyperion~\cite{10.1145/3460319.3464811}, but without requiring manually defined features. By constructing maps of the input space, \topomap enables developers to visualise how inputs are organised according to their features and to identify regions that are under-represented in the training data. Concentrating testing efforts on clusters populated by fault-revealing features enables the creation of targeted and challenging test inputs, improving fault detection without requiring exhaustive exploration of the entire input space. In addition, the spatial organisation of clusters can be analysed to identify groups of neighbouring regions that collectively expose faults, %failures
revealing larger macro-regions of vulnerability. Finally, once constructed, a \topomap can serve as a reusable mapping of the input space. As new test inputs become available, they can be projected onto the existing map allowing practitioners to analyse the changes in the input space coverage, localisation and connections between failure-inducing inputs. Additionally, it may serve as a visualisation tool to analyse changes in the model behaviour once the model is retrained or changed. This makes the approach naturally suited to iterative testing and continuous integration workflows.

%\gianmarco{Add mentions of DeepGini and Surprise Adequacy, probably add a Discussion section. They achieve better results, but they are white-box.} 
While white-box techniques such as DeepGini~\cite{feng2020deepgini} and Surprise Adequacy~\cite{kim2019guiding, 11026894} were proved effective at finding and prioritising critical inputs, they partition the input space based on the inner model behaviour (the model's confidence or activations, respectively). We believe the two perspectives (black vs white box) on the problem of input partitioning for DL testing are complementary and developers can benefit from both. We conjecture that by combining them we can achieve higher diversity and spread of the misbehaviour exposing inputs.
%It is expected that the aforementioned techniques, by using information obtained from the model, e.g., the output probabilities or latent space representations of the inputs, may be more effective in prioritisation and identification of critical inputs, but their applicability is bound to the access level of the model under test. Furthermore, both kinds of techniques do not insert test inputs into a broader relational context with respect to the dataset feature space, thus having a high risk of ``over-exploring" specific regions of the feature space.   

%\gianmarco{Some operators may not be localised, then some tend to be localised. Some operators that may sound global, they may be localised depending on the easiness of the input features.}
%Our analysis evaluates the ability of test inputs in specific regions to trigger misbehaviour in DeepCrime mutants. This draws a link between some specific features and model faults. 

A crucial aspect of \topomap lies in the trade-off between specificity and generalisability. While the framework itself is adaptable to various scenarios, the usage of specific embedding and clustering algorithms affects the quality of the produced topographical map. In particular, in this experimental analysis, we leverage low-level features of the raw input space enhanced by basic linear and non-linear embedding algorithms. This allows us to extend the applicability of our empirical pipeline to datasets pertaining to different domains, including images, speech, and text. In concrete application scenarios, using domain-specific embeddings leads to neater separation between feature clusters, possibly powered by semantic meaning. 
Moreover, \topomap relies on the capability of a DNN to learn the boundaries between regions characterised by different features in the input space. While DNNs are general function approximators that can in principle learn such boundaries, we acknowledge that  semantically meaningful and discriminative regions may exist, which are difficult to learn for a DNN. Hence, rather than claiming any intrinsic discriminative capability of \topomap, we refer to the empirical validation of the identified regions to justify the usage of a DNN as a practical way to find relevant boundaries.

\section{Conclusion}\label{sec:concl}
In this work, we propose \topomap, a model-agnostic approach that generates a topographical map of the DNN input feature space. We apply this approach to explore and identify the map regions that share features that are likely to expose faults in a DL model. We also propose an automated technique to configure the dimensionality reduction and clustering methods used in our approach, to optimise the discriminative power of the resulting map regions. We evaluated the effectiveness of \topomap in selecting fault-exposing %failure-exposing
inputs. The results indicate that in terms of killing probability, our approach outperforms the random baseline by 35\%\textit{pt} on average on killable mutants and by 74\%\textit{pt} on initially non-killable ones. Compared to DeepHyperion on the MNIST subject, \topomap has a higher killing contributor density, increased by 8\%\textit{pt} on killable mutants and 12\%\textit{pt} on initially non-killable mutants.

Beyond quantitative performance, our analysis shows that the maps produced by \topomap highlight regions that are both meaningful and easily distinguishable. Importantly, our human study indicates that participants’ judgements tend to align with the input grouping of \topomap, particularly when the images in the dataset are straightforward to interpret. This supports the interpretability and practical usage of the generated maps.

By focussing their testing effort on map regions that expose misbehaviours of the model under test, developers can efficiently create a challenging test set for the faults possibly affecting their model. In our future work, our aim is to rely on the features exposed by \topomap to generate new input instances tailored to target specific failure-rich 
regions of the map when testing DL systems. Furthermore, it may be interesting to exploit the spatial relations between killing clusters, to identify potential macro-regions of contributing inputs.

\section*{Data Availability}

All data, scripts, and results of this study are publicly available~\cite{github:repo}.

\section*{Ethics approval}

The human study was approved by the Ethics Committee of the Università della Svizzera italiana (Decision CE-2025-14).

\section*{Funding}
This work is funded by the Swiss National Science Foundation (SNSF) programme under the project Toposcope, grant agreement n. 214989.

\bibliographystyle{plainnat}

\bibliography{sample-base}   % name your BibTeX data base

\end{document}